\documentclass[a4paper,fleqn]{cas-dc}

\usepackage[numbers]{natbib}
\usepackage{bm}

\ExplSyntaxOn
\cs_set:Npn \__first_footerline: {}
\ExplSyntaxOff

\begin{document}
\let\WriteBookmarks\relax
\def\floatpagepagefraction{1}
\def\textpagefraction{.001}
\shorttitle{FGD2HF}

\title [mode = title]{FGD2HF: Frequency-guided Dual-domain Hypergraph Fusion for Semantic Segmentation}                      

\author[1]{Xiaofan Yang}[style=chinese,
                        bioid=1,
                        orcid=0009-0007-0406-1096]

\affiliation[1]{organization={Harbin Institute of Technology}, 
            addressline={Xidazhi Street 92, Nangang Dist.}, 
            city={Harbin},
            postcode={150001}, 
            state={Heilongjiang},
            country={China}}

\author[2]{Wei Pan}[style=chinese]

\affiliation[2]{organization={Harbin Engineering University}, 
            addressline={Nantong Street 145, Nangang Dist.}, 
            city={Harbin},
            postcode={150001}, 
            state={Heilongjiang},
            country={China}}

\author[1]{Yubin Liu}[style=chinese]

\cormark[1]

\author[2]{Junyu Wu}[style=chinese]

\author[1]{Junming Zhang}[style=chinese]

\cormark[1]

\author[1]{Jie Zhao}[style=chinese]

\begin{abstract}
Frequency-domain information is beneficial for semantic segmentation. However, existing frequency-based methods often suffer from insufficient frequency utilization, inefficient fusion, and high computational complexity. To address these issues, we propose FGD2HF, a dual-domain fusion network that leverages frequency features to guide spatial-domain fusion and enhancement. Specifically, we introduce a novel Hybrid-Attention Encoder (NAHA) that jointly models local and global relationships through neighborhood and hypergraph attention, enhancing the network's ability to extract effective features. We further introduce a Wavelet Transform and Enhancement (WTE) module to extract and refine low- and high-frequency cues from preceding-stage features, followed by Frequency-guided Dual-domain Fusion (FGDF) modules that use frequency cues to condition current-stage spatial grouping and feature propagation. This design incorporates both global and local frequency guidance into spatial feature modulation, enhancement and fusion with low computational cost. Extensive comparisons with numerous state-of-the-art semantic segmentation methods demonstrate that our method is highly competitive. Our full FGD2HF-B model achieves 48.7\% mIoU on ADE20K (+0.3\% over SegMAN-B with 18\% fewer GFLOPs), 83.2\% mIoU on Cityscapes (comparable to SegMAN-B with 21\% fewer GFLOPs), and 1.1\% higher mIoU than VWFormer-T on SBD with 21\% fewer GFLOPs, while also achieving significant improvements in boundary segmentation accuracy. The code will be made publicly available upon paper acceptance.
\end{abstract}


\maketitle

\section{Introduction}

Semantic segmentation, as an important downstream task in computer vision, has been widely applied in various fields, including autonomous driving \cite{li2026dyfuse, wang2025reliable, silva2025exploring}, medical diagnosis \cite{yang2025hsrdiff, krithika2022review}, satellite image analysis \cite{singh2023semantic, amirgan2024semantic}, and robotic operations \cite{zhong2024novel, brunke2025semantically, jiang2024geometry}. In recent years, with the rapid development of deep learning and convolutional neural networks, CNN-based segmentation models such as Fully Convolutional Networks (FCN) \cite{long2015fully}, MASK R-CNN \cite{he2017mask}, U-Net \cite{ronneberger2015u}, and DeepLab \cite{chen2017deeplab} have gradually replaced traditional semantic segmentation methods and become the mainstream research paradigm. Moreover, since the breakthrough of the visual attention mechanism represented by Vision Transformer (ViT) \cite{dosovitskiy2020image}, Transformer-based semantic segmentation models \cite{zheng2021rethinking, strudel2021segmenter, xie2021segformer, chen2021dpt, cheng2021per, cheng2022masked} have experienced rapid development. Compared with conventional CNN-based models, Transformer-based approaches benefit from global receptive fields and strong spatial relationship modeling capabilities, thereby significantly improving segmentation accuracy \cite{thisanke2023semantic}.
\begin{figure}
    \centering
    \includegraphics[width=\columnwidth]{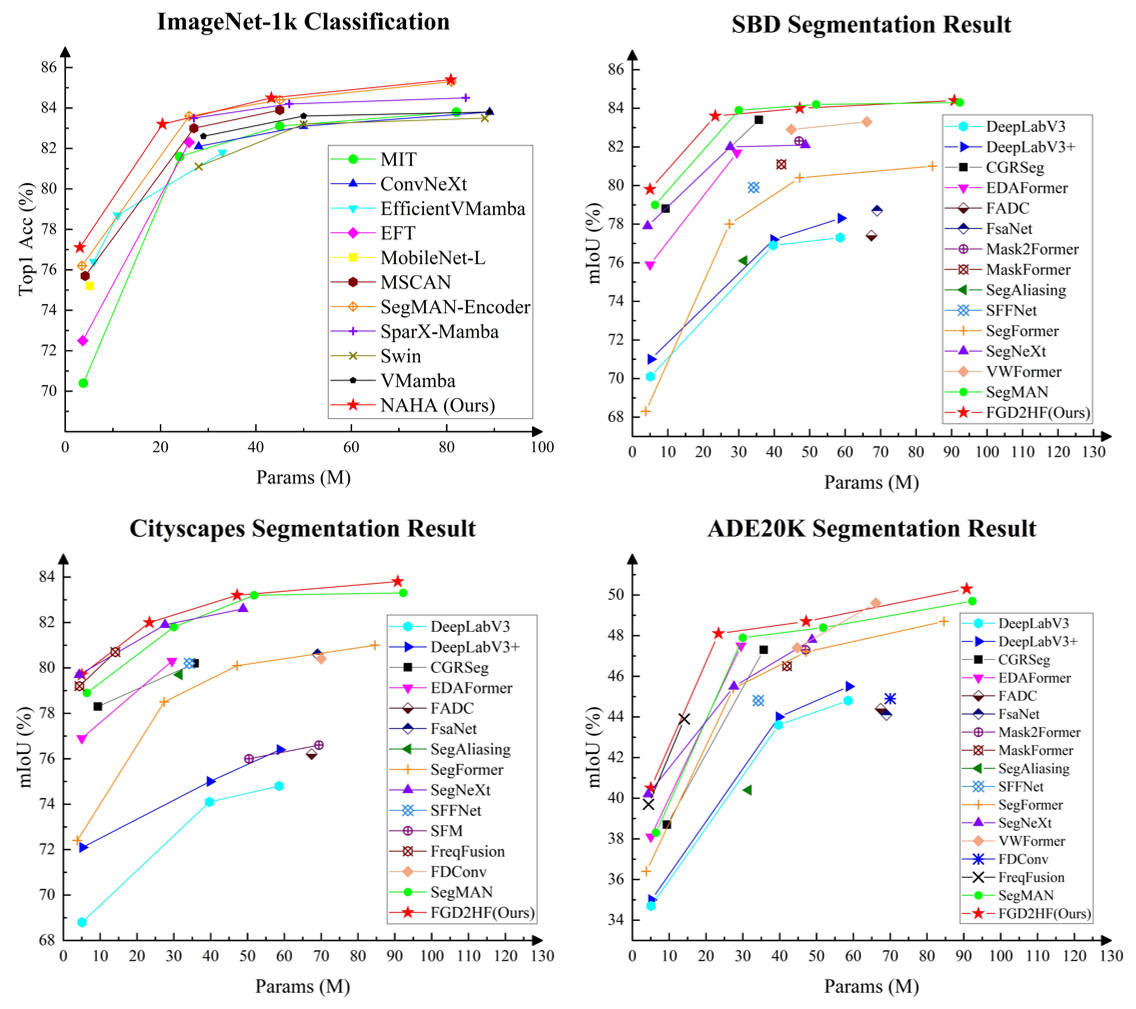}
    \caption{NAHA Encoder classification performance compared with representative feature extraction backbones and semantic segmentation results of FGD2HF compared to state-of-the-art methods.}
    \label{fig1}
\end{figure}

However, traditional semantic segmentation models that rely solely on spatial information have inherent limitations. While spatial features are intuitive and effective, they are not sufficiently robust and are easily affected by noise, illumination changes, and distribution shifts \cite{bi2024learning}. Furthermore, they often perform poorly in modeling object boundaries and capturing subtle inter-class differences. In contrast, frequency-domain features offer unique advantages in global representation and resistance to interference \cite{aslan2026frequency}. First, global spectral transforms such as Fourier and DCT provide holistic frequency representations, whereas wavelet transforms jointly preserve spatial localization and multi-scale frequency characteristics. Second, frequency-domain decomposition can effectively separate low- and high-frequency components. High-frequency components capture object contours, edges, and fine details, while low-frequency components encode global illumination, color distribution, and semantic structures \cite{li2025frequency}.  These properties are particularly valuable for semantic segmentation as a dense prediction task.

Therefore, effectively leveraging frequency-domain features to improve segmentation accuracy has gradually become an important research direction in recent years. In contrast to spatial-domain approaches, frequency-domain features capture different types of image information from a signal distribution perspective. By incorporating frequency-domain feature extraction, selection, and fusion, segmentation models can effectively suppress semantically irrelevant noise while reinforcing consistent target representations. This approach significantly enhances both segmentation accuracy and robustness in complex scenes.

We categorize existing frequency-based semantic segmentation methods into three technical routes. The first route focuses primarily on low-frequency information to enhance the consistency of target regions and global structural representations. Representative methods include DCT-EDANet \cite{lo2019exploring}, FDA \cite{yang2020fda}, and FsaNet \cite{zhang2023fsanet}. The second route further exploits the distinction between low- and high-frequency components. By decoupling, selecting, calibrating, or adaptively fusing low- and high-frequency components, these methods enhance boundary and detail preservation while maintaining global context consistency. Representative methods include FDNet \cite{li2025frequency}, FADA \cite{bi2024learning}, FADC \cite{chen2024frequency}, and FFR \cite{yang2025ffr}. The third route underscores the importance of dual-domain collaborative modeling, facilitating seamless interaction between spatial and frequency representations. By integrating low- and high-frequency information with spatial features, these methods achieve improved segmentation performance in complex scenes and areas with ambiguous boundaries. Representative methods include DDFNet \cite{li2025dual} and SFFNet \cite{yang2024sffnet}. Compared with the first two routes, dual-domain methods fully exploit the complementary advantages of frequency and spatial features, demonstrating stronger adaptability in complex scenarios.

However, existing spatial-frequency collaborative methods still face challenges. First, dual-domain feature extraction and fusion often require additional transformations, feature projections, and multi-branch interactions, which increase computational cost and memory consumption, especially for high-resolution images. Second, spatial and frequency features differ significantly in terms of representation form, semantic level, and responsive regions. Simple operations such as concatenation, addition, or weighted fusion may not achieve effective alignment. Instead, they risk introducing redundant information or semantic conflicts. Third, CNNs and Transformers tend to exhibit a low-pass filtering effect during deep feature extraction, where downsampling and global modeling progressively weaken high-frequency components, resulting in blurred boundaries, broken fine structures, and incomplete object contours \cite{thisanke2023semantic}. Therefore, achieving effective fusion between spatial and frequency features with low computational cost, while simultaneously preserving high-frequency boundary information, remains a key challenge in frequency-based semantic segmentation.
 
To address these challenges, we propose a dual-domain feature fusion framework named FGD2HF for semantic segmentation that systematically addresses three key aspects: (1) computational efficiency, (2) efficient cross-domain feature fusion, and (3) precise boundary preservation. FGD2HF employs Haar decomposition to decouple low- and high-frequency features from the preceding stage and uses the resulting frequency subbands to guide current-stage spatial feature enhancement and dual-domain fusion. The spatial locality of Haar decomposition and its downsampling factor establish a structured region-wise correspondence with current-stage features, alleviating the alignment burden in network design and addressing the problem of high-frequency propagation. Specifically, we introduce an Encoder called NAHA that jointly models local and global relationships through neighborhood and hypergraph attention, enhancing the network's ability to extract effective features. We further introduce a Wavelet Transform and Enhancement (WTE) module to extract and refine low- and high-frequency cues from preceding-stage features, followed by Frequency-guided Dual-domain Fusion (FGDF) modules that use frequency cues to condition current-stage spatial grouping and feature propagation. Its frequency-guided hypergraph attention organizes spatial nodes into overlapping groups, with soft memberships jointly determined by global frequency context and aligned local frequency cues. Compared with standard full self-attention over ($N$) spatial tokens, using ($M$) hyperedges, where ($M \ll N$), reduces the number of relation weights from ($N^2$) to ($NM$), enabling compact spatial-frequency interaction.

In summary, our contributions are as follows:

(1) We propose FGD2HF, a Frequency-guided Dual-domain Hypergraph Fusion network for semantic segmentation. The framework decomposes preceding-stage features into low- and high-frequency subbands and adaptively uses the resulting frequency cues to guide current-stage spatial feature enhancement and fusion, boosting both segmentation accuracy and boundary precision.

(2) We develop a novel Hybrid-Attention Encoder, termed NAHA, which integrates neighborhood attention and hypergraph attention to jointly capture local and global dependencies, enhancing the network's feature extraction capability while reducing computational complexity.

(3) We design a Wavelet Transform and Enhancement (WTE) module for high- and low-frequency decomposition and enhancement, and a Frequency-guided Dual-domain Fusion (FGDF) module that leverages frequency information to modulate spatial information, enabling efficient dual-domain feature fusion at low computational cost.

(4) Extensive experiments on multiple benchmark datasets demonstrate that our FGD2HF achieves highly competitive performance across diverse scenarios. Comprehensive ablation studies further verify the effectiveness and superiority of each proposed component.

\section{Related Works}
\subsection{Semantic Segmentation}
Semantic segmentation is a dense prediction task in computer vision that aims to assign a label to every pixel in an image \cite{fu2025segman}. It is widely applied in autonomous driving, robotic perception, medical analysis, and remote sensing. Early semantic segmentation approaches primarily relied on hand-crafted, low-level visual features such as edges, colors, textures, and region similarity \cite{csurka2022semantic}. Representative methods include edge detection techniques \cite{arbelaez2010contour, wang2010generalizing}, clustering-based algorithms \cite{comaniciu2002mean, achanta2012slic}, support vector machines \cite{pal2005support, wang2011color}, and Markov random fields \cite{geman1984stochastic, shotton2009textonboost}. Although effective in simple scenarios, these methods often lack the representation capacity needed for accurate and robust segmentation in complex scenes with intricate structures, illumination changes, and noise.

With the rapid advancement of deep learning, CNN-based models \cite{long2015fully, ronneberger2015u, badrinarayanan2017segnet, chen2017deeplab} have gradually become mainstream. These methods significantly improve segmentation performance by learning hierarchical spatial features in an end-to-end manner. However, CNNs primarily rely on local convolution operations \cite{younesi2024comprehensive}. While effective at modeling local textures, their restricted receptive fields limit their ability to capture global contextual information. The introduction of attention mechanisms spurred the development of Transformer-based segmentation models. Early methods such as SETR \cite{zheng2021rethinking} and Segmenter \cite{strudel2021segmenter} employ ViT as the backbone for feature extraction, verifying the feasibility of Transformer architectures in dense prediction tasks. However, these approaches incur substantial computational overhead because the self-attention mechanism scales quadratically with the number of input tokens. Subsequently, methods such as Swin Transformer \cite{liu2021swin}, PVT \cite{wang2021pyramid}, and SegFormer \cite{xie2021segformer} introduced hierarchical structures and multi-scale features to address the lack of local inductive bias and the high computational overhead in ViT. Furthermore, MaskFormer \cite{cheng2021per} and Mask2Former \cite{cheng2022masked} transformed the segmentation task from pixel-by-pixel classification to a mask classification problem, achieving unified modeling of semantic segmentation, instance segmentation, and panoptic segmentation. However, these methods mainly rely on spatial features and correlations, without explicitly modeling frequency characteristics or boundary details.

Given that frequency features are effective at capturing global structures, texture details, and boundary variations, using frequency information has become an important trend in semantic segmentation research. Early studies primarily focused on low-frequency and global information. For example, DCT-EDANet directly utilizes DCT compressed-domain representations for segmentation \cite{lo2019exploring}. FsaNet reduces computational complexity and enhances intra-object consistency through low-frequency self-attention \cite{zhang2023fsanet}. FDA achieves domain-style alignment using Fourier low-frequency components \cite{yang2020fda}. However, relying solely on low-frequency information limits performance in complex backgrounds, sharp boundaries, and subtle inter-class texture variations. This has motivated the joint optimization of low- and high-frequency features. For instance, FADA decomposes basic visual features into low- and high-frequency components using Haar wavelets to improve domain generalization \cite{bi2024learning}. FADC adaptively adjusts the receptive field and frequency response of atrous convolution from a spectral perspective \cite{chen2024frequency}. FFR corrects the boundary over-smoothing problem caused by high-frequency attenuation in ViT-based weakly supervised segmentation \cite{yang2025ffr}. These methods improve boundary preservation and detail recovery, but largely model frequency features in isolation, leaving their complementarity with spatial semantics underexplored.

More recently, advanced methods have begun to explore deep cross-domain interactions between frequency components and spatial features. DDFNet \cite{li2025dual} selectively integrates low- and high-frequency components from both domains through a dynamic decoupling and fusion strategy, enabling the simultaneous capture of local textures and global semantics. SFFNet \cite{yang2024sffnet} decomposes features using Haar wavelets and then aligns and fuses low-frequency features with spatial global semantics and high-frequency features with spatial local details, thereby achieving effective complementarity between the frequency and spatial domains.

\subsection{Frequency Domain Feature Learning}
Frequency-domain analysis, as a powerful signal processing tool, provides a new perspective for understanding model optimization and generalization. In recent years, frequency-domain learning has been widely applied to address issues such as spectral bias, detail degradation, and artifacts in neural networks, achieving remarkable progress in image classification, object detection, and semantic segmentation \cite{stuchi2024frequency, hou2022multisignal, ling2025fdl, aslan2026frequency}. Kai Xu et al. introduced a DCT-based frequency-domain learning framework that reduces input redundancy while maintaining or improving recognition accuracy \cite{xu2020learning}. Yijie Zhong et al. incorporated frequency cues beyond RGB information to enhance camouflaged object detection \cite{zhong2022detecting}. FreqNet forces detectors to learn frequency-domain representations rather than relying on forgery-specific artifacts, improving generalization to unseen deepfake sources \cite{tan2024frequency}.

In semantic segmentation, FsaNet \cite{zhang2023fsanet} introduces a frequency-domain attention mechanism that performs attention on low-frequency components, reducing computational complexity while maintaining segmentation performance. FADC \cite{chen2024frequency} mitigates high-frequency detail loss and gridding artifacts in dilated convolution by adapting dilation rates to local high-frequency energy. FFR \cite{yang2025ffr} addresses the low-pass bias of Vision Transformers with a dual-branch head that separately processes original and high-frequency features for sharper boundaries. FDNet \cite{li2025frequency} decouples low- and high-frequency components using learnable Gaussian kernels, recalibrates frequency channels, and applies frequency attention for fusion. DDFNet \cite{li2025dual} further performs dynamic cross-domain decoupling and fusion to jointly model local textures and global semantics.

\subsection{Hypergraph Neural Network}
Hypergraphs are an extension of conventional graphs in graph theory and provide a more flexible framework for modeling multi-way relationships \cite{bretto2013hypergraph}. Unlike ordinary graphs, where each edge connects only two vertices, a hypergraph introduces the concept of hyperedges, each of which can connect an arbitrary number of vertices. In contrast to self-attention, which models all pairwise token interactions, hypergraph modeling captures global high-order dependencies through a small set of adaptive hyperedges, achieving near-linear complexity. This enables hypergraphs to capture global many-to-many relationships, making them well suited to complex relational tasks such as clustering, object detection, and recommendation systems \cite{li2023efficient, feng2023modularity, guo2025mmhcl, khan2025heterogeneous}. Eun-Sol Kim et al. proposed Hypergraph Attention Networks (HANs), a multimodal learning method designed to address the inconsistency of information hierarchies across different modalities, such as images and text \cite{kim2020hypergraph}. HHGSA \cite{khan2025heterogeneous} leverages heterogeneous hypergraphs and attention mechanisms to model high-order social relationships and item similarities, significantly improving recommendation accuracy. YOLOv13 \cite{lei2025yolov13} introduces hypergraph mechanism to object detection, effectively boosting detection performance. M2I2HA \cite{yang2026m2i2ha} extends hypergraph attention to multimodal detection by constructing cross-modal hypergraphs and employing low-rank decomposition to reduce complexity, substantially improving detection accuracy.

In semantic segmentation, Shurong Chai et al. proposed an adaptive hypergraph neural network (AHGNN) for medical image segmentation, aiming to overcome the limitations of Transformers and graph convolutional networks in modeling non-pairwise relationships \cite{chai2024novel}. Weipeng Jing et al. proposed Hypergraph BiFormer (HGBT), a novel architecture for semantic segmentation of high-resolution remote sensing images. By introducing hypergraph structures, HGBT enhances the ability to model local features and segment multi-scale objects \cite{jing2025hypergraph}. Manothini Palanisamy et al. integrated hypergraph neural networks with multi-scale relational encoding for prostate cancer diagnosis. This framework combines spatial and spectral attention to improve boundary segmentation accuracy, while incorporating adaptive feature selection to enhance model generalization \cite{palanisamy2026multi}. While hypergraph theory has already found applications in semantic segmentation, cross-domain relational modeling between the spatial and frequency domains based on hypergraphs remains largely unexplored. This underexplored direction holds great promise and calls for further investigation.

\section{Methodology}
In this section, we provide a detailed introduction to the proposed method. We begin by presenting our observations on image frequency decomposition and the design motivations. Then, we present an overview of the architecture, followed by in-depth discussions of the NAHA Encoder, the WTE module, the FGDF module, the Decoder, and the loss function design.
\subsection{Preliminaries}
\textbf{Hypergraph}. Graph structures excel at modeling pairwise relationships, but their descriptive capacity becomes limited when confronted with the multi-way associations prevalent in real-world scenarios. As a generalization of conventional graphs, hypergraphs allow hyperedges to connect any number of nodes, thereby offering an intuitive and flexible means to represent high-order or group-wise relationships among entities.

A hypergraph can be formally represented as $\mathcal{W} = (\mathcal{V}, \mathcal{E})$, where $\mathcal{V} = \{v_1,v_2,...,v_n\}$ denotes a finite set of $n$ nodes, and $\mathcal{E} = \{e_1,e_2,...,e_m\}$ is a collection of $m$ hyperedges, each of which is a non empty subset of $\mathcal{V}$ (i.e., $e \subseteq  \mathcal{V}$ and $e \neq  \varnothing $). The structure of a hypergraph is commonly described by an incidence matrix $\bm{A}  \in \mathbb{R} ^{n\times m}$, whose elements are defined as follows:
\begin{equation}
\bm{A}_{i,j} =
\begin{cases}
1, & \text{if vertex } v_i \in \text{hyperedge } e_j \\
0, & \text{otherwise}
\end{cases}
\end{equation}

\textbf{Wavelet Transformation}. Wavelet transformation is a signal analysis technique that jointly characterizes spatial and frequency information. For a 2D image or feature map, the wavelet transform applies low-pass and high-pass filtering along the horizontal and vertical directions, followed by downsampling, thereby decomposing the input into four frequency sub-bands: (1) the low-frequency component $LL$, (2) the high-frequency components $LH$, $HL$, and $HH$, which capture horizontal, vertical, and diagonal variations, respectively.

Given an input feature map $\bm{F} \in \mathbb{R}^{H\times W\times C}$, a one-level 2D discrete wavelet transform can be formulated as
\begin{equation}
  \left\{ LL, LH, HL, HH \right\} = \operatorname{WT}(\bm{F})
\end{equation}
where each sub-band generally has a spatial resolution of $\frac{H}{2}\times \frac{W}{2}$. Therefore, wavelet decomposition explicitly separates structural information from fine-grained details while reducing the resolution. WT($\cdot$) denotes the 2D discrete wavelet transform operation. We can also reconstruct the decomposed frequency sub-bands to the original map without any loss by inverse wavelet transform IWT($\cdot$), i.e.
\begin{equation}
    \bm{F} = \operatorname{IWT}(LL, LH, HL, HH)
\end{equation}

Compared with conventional pooling or strided convolution, wavelet decomposition provides a fixed and interpretable frequency decomposition process and can alleviate the loss of detailed information during downsampling. Among different wavelet bases, the Haar wavelet is particularly attractive for lightweight vision networks because of its simple formulation and high computational efficiency.

\subsection{Observations and Motivations} 
As illustrated in Fig. 2(a)-(d), we apply Haar wavelet decomposition to an example image and reconstruct its low- and high-frequency components separately. The high-frequency reconstruction is enhanced and normalized for visualization only. The low-frequency reconstruction retains the overall appearance and coarse structure, with some blurring around object boundaries, whereas the high-frequency reconstruction emphasizes edges and fine texture. For this example, the low- and high-frequency components account for 98.65\% and 1.35\% of the total coefficient energy, respectively. Reconstruction using all four subbands recovers the analyzed input. 

To assess whether this observation extends beyond the example, we compute dataset-level statistics, as shown in Fig. 2(e)-(h). The mean low-frequency energy ratios are 98.68\%, 99.08\%, and 99.83\% on SBD, ADE20K, and Cityscapes, respectively, with corresponding high-frequency ratios of 1.315\%, 0.922\%, and 0.173\%. These results indicate that low-frequency components dominate the average signal energy across the three analyzed datasets under the stated preprocessing protocol, while the precise ratios vary. Although energy magnitude alone does not establish a component's importance for segmentation, the complementary structures visible in the reconstructions motivate incorporating both low-frequency context and high-frequency detail into spatial feature enhancement.

\begin{figure}
\centering
\includegraphics[width=\columnwidth]{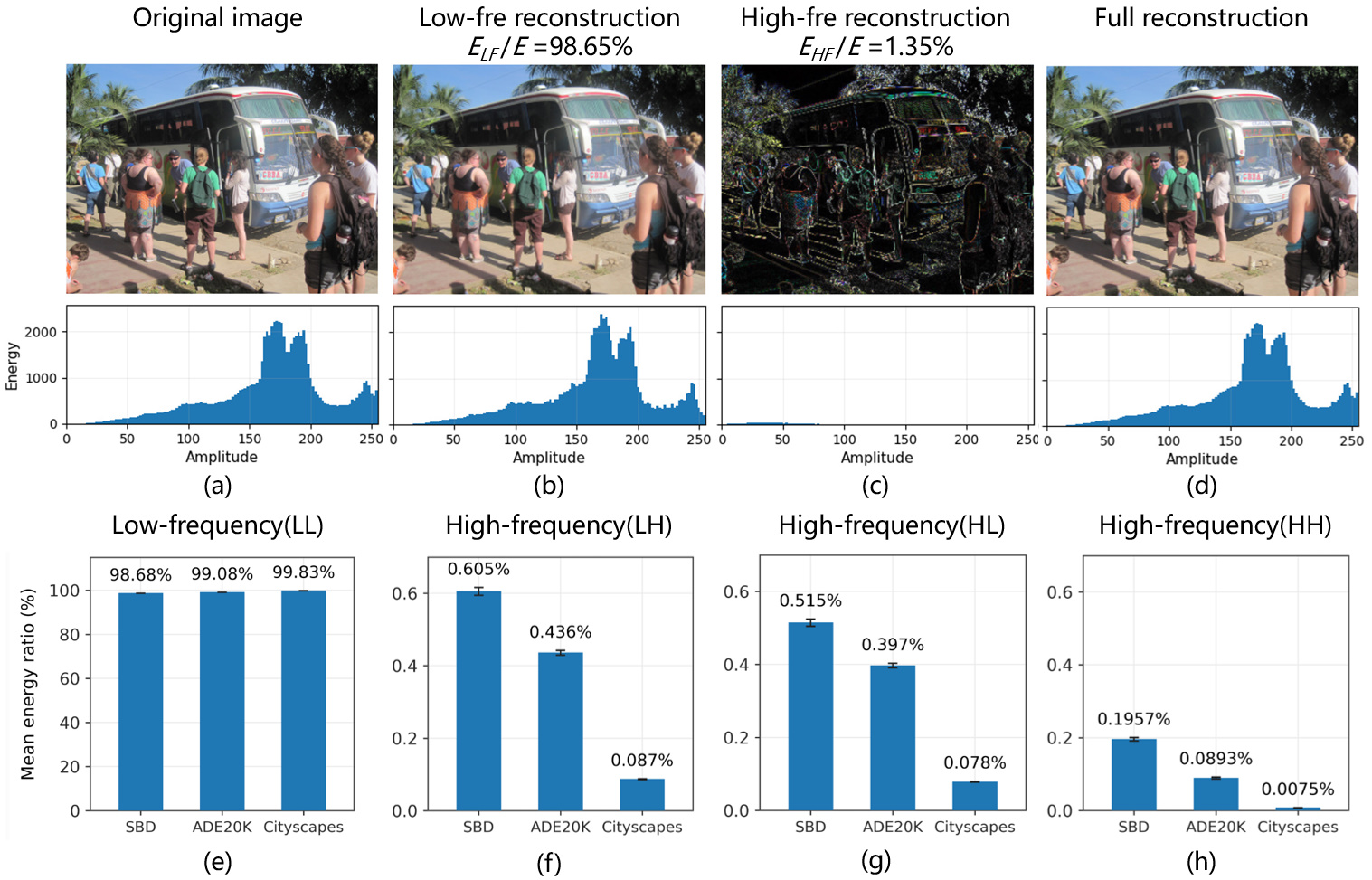}
\caption{Visualization of wavelet decomposition, image reconstruction and dataset-level energy statistics. (a) Original example image. (b) Low-frequency reconstruction. (c) High-frequency reconstruction. (d) Full reconstruction. Energy percentages in (b,c) are for the example only and computed before enhancement. (e-h) Mean LL/LH/HL/HH energy ratios on different datasets. The energy ratio is computed from the squared $L_2$-norm of the Haar wavelet coefficients.}
\label{fig2}
\end{figure}

Based on these observations, we formulate the following design motivations:

(1) \textbf{High-frequency components encode critical boundary cues.} However, neural networks generally exhibit a spectral bias toward low-frequency information, limiting their ability to capture fine-grained details \cite{rahaman2019spectral, xu2019frequency, magid2021dynamic}. Moreover, attention layers exacerbate this issue \cite{li2025frequency}. \textbf{Therefore, we can propagate high-frequency features from shallow stages to deeper stages for frequency-aware enhancement and fusion, improving boundary preservation and feature representation.}

(2) Incorporating frequency components inevitably introduces additional computational overhead, especially in attention-based networks. Haar wavelet decomposition of preceding-stage spatial features produces spatially localized frequency subbands at the resolution of the current stage. These subbands support region-wise correspondence with current-stage spatial features. \textbf{Thus, decomposing features from the preceding-stage to enhance the current-stage not only supports the first motivation but also naturally aligns the resolutions of adjacent stages, facilitating spatial-frequency fusion and easing the alignment burden that may arise when fusing independently processed dual-domain branches.}

(3) High-frequency components may contain both boundary-relevant details and task-irrelevant noise. \textbf{We therefore introduce an auxiliary boundary prediction task on the high-frequency-guided spatial features. This supervision encourages these representations to encode discriminative boundary cues, complementing the main semantic segmentation objective.}

\begin{figure*}
    \centering
    \includegraphics[width=\textwidth]{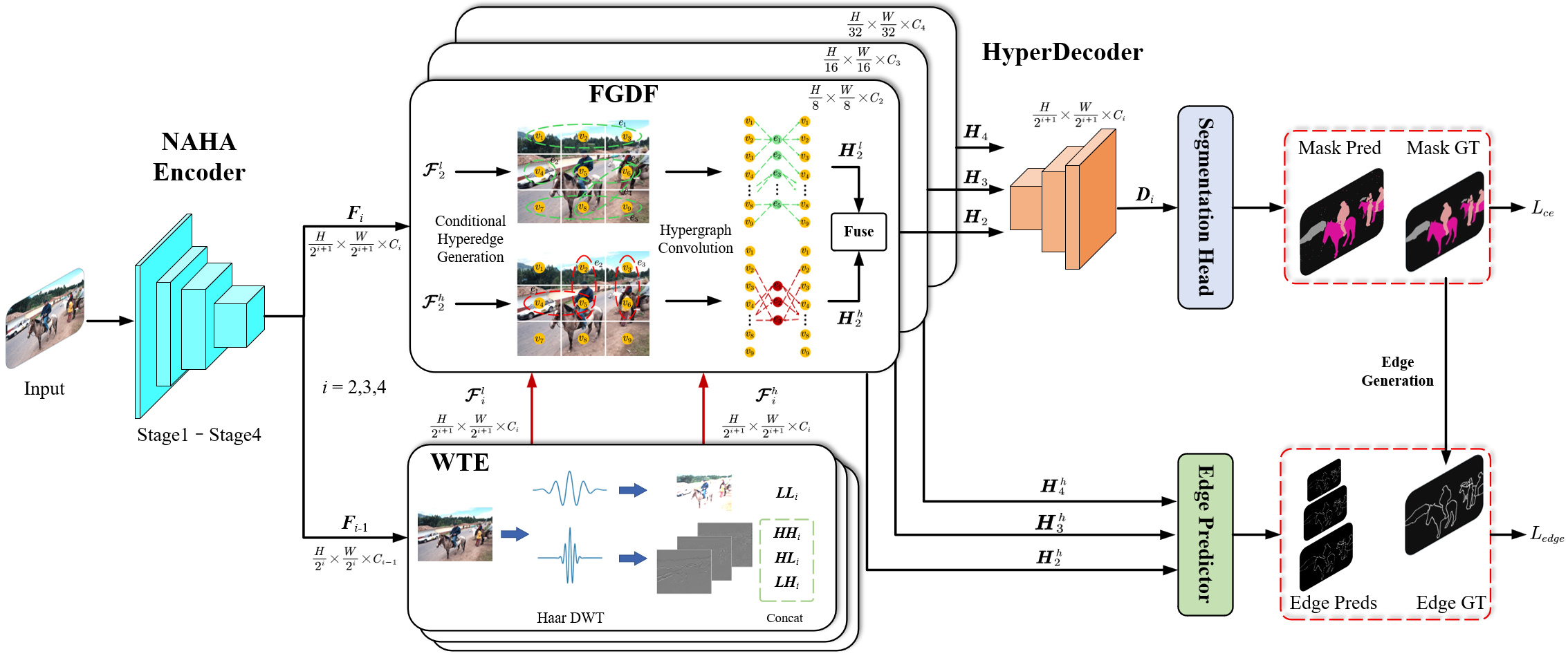}
    \caption{Overall architecture of the proposed FGD2HF network.}
    \label{fig3}
\end{figure*}

\subsection{Overall Architecture} 
Existing spatial-frequency fusion methods often process spatial and frequency features in parallel before fusion, which can introduce additional computational overhead and complicate cross-domain feature alignment. In contrast, our framework applies Haar wavelet decomposition directly to the encoded features of the preceding stage. Wavelet decomposition's spatial locality, combined with its downsampling factor matching the inter-stage resolution change, enables region-wise correspondence between its subbands and current-stage spatial features \cite{han2026fdhnet}. This design reuses the existing feature hierarchy for frequency extraction, reducing the overhead associated with independent frequency encoding. The resulting frequency cues guide the grouping and enhancement of current-stage spatial features through hypergraph theory, achieving adaptive spatial-frequency fusion and fine-grained representation. Moreover, the enhanced features from the high-frequency-guided branch support multi-scale auxiliary boundary prediction, which encourages adaptive use of boundary-relevant high-frequency cues and mitigates boundary-detail loss caused by high-frequency attenuation in deep networks.

Fig.~\ref{fig3} illustrates the overall architecture of the proposed FGD2HF network, which comprises a hybrid-attention encoder (NAHA), a Wavelet Transform and Enhancement (WTE) module, Frequency-guided Dual-domain Fusion (FGDF) modules, and a HyperDecoder. The main processing stages are described as follows. 

(1) During encoding, NAHA combines neighborhood attention and hypergraph attention to extract multi-scale features $\{\bm{F}_i\}_{i=1}^{4}$. Neighborhood attention captures local relationships, while hypergraph attention models global high-order interactions, enabling complementary local and global feature learning.

(2) For each stage $i\in\{2,3,4\}$, WTE applies Haar wavelet decomposition to the preceding-stage feature $\bm{F}_{i-1}$, yielding four spatially localized subbands $\{LL_i,LH_i,HL_i, \allowbreak HH_i\}$ at the resolution of the current-stage feature $\bm{F}_i$. The three high-frequency subbands are concatenated along the channel dimension, after which low-frequency refinement and low-frequency-conditioned high-frequency modulation are applied to obtain $\mathcal{F}_i^{l}$ and $\mathcal{F}_i^{h}$.

(3) Each FGDF module uses $\mathcal{F}_i^{l}$ and $\mathcal{F}_i^{h}$ to guide spatial feature enhancement through separate low- and high-frequency branches. Within each branch, global frequency context and aligned local frequency cues jointly condition node--hyperedge associations. Spatial features are aggregated into a compact set of hyperedges, transformed nonlinearly, and propagated back to the nodes, enabling frequency-conditioned high-order feature enhancement and spatial--frequency fusion. The two branch outputs $\bm{H}_i^{l}$ and $\bm{H}_i^{h}$ are concatenated and fused by a convolution, followed by a scaled residual connection with $\bm{F}_i$ to produce $\bm{H}_i$.

(4) During decoding, the HyperDecoder further refines the enhanced multi-scale features $\{\bm{H}_i\}_{i=2}^{4}$ and obtains the decoder representations $\{\bm{D}_i\}_{i=2}^{4}$.

(5) Finally, the decoder representations are fed into a segmentation head to generate the dense prediction, which is supervised by a pixel-wise cross-entropy loss. Meanwhile, the high-frequency-guided features $\{\bm{H}_i^{h}\}_{i=2}^{4}$ are passed to an edge predictor to generate multi-scale boundary predictions, providing deep supervision through an auxiliary boundary loss.

\subsection{NAHA Encoder} 
\begin{figure*}
    \centering
    \includegraphics[width=0.9\textwidth]{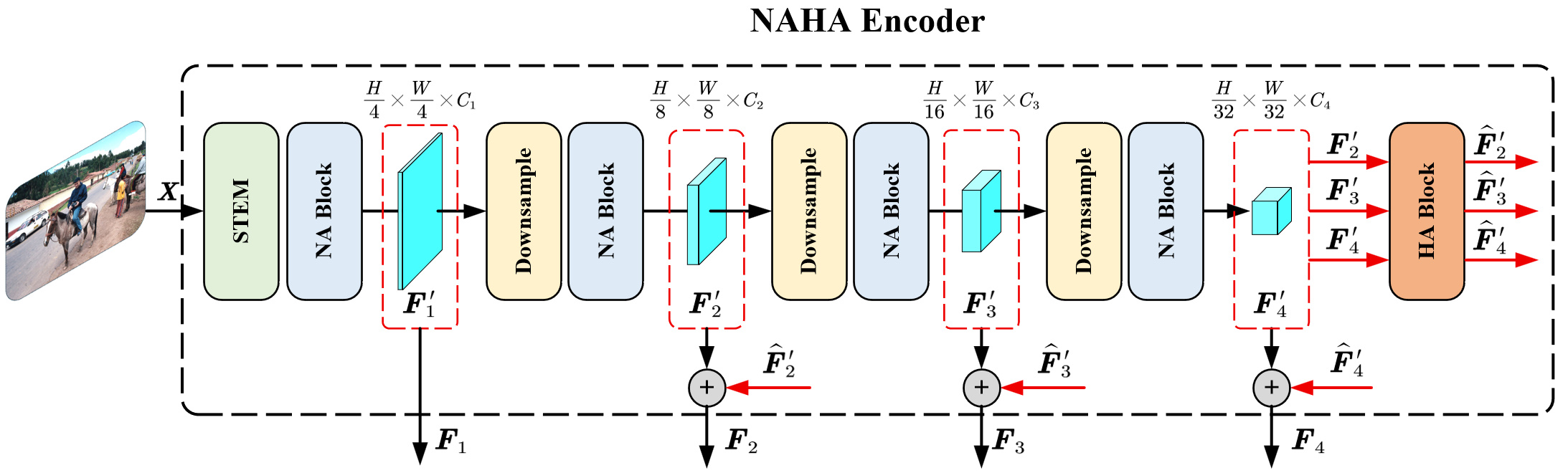}
    \caption{The structure of the NAHA encoder.}
    \label{fignaha}
\end{figure*}

Previous studies have shown that local context, global context, and multi-scale context modeling across intermediate features are fundamental to accurate semantic segmentation \cite{yu2018learning, fu2019dual, fu2025segman}. To enhance the network's feature extraction capability and provide high-quality semantic representations for subsequent frequency feature extraction and dual-domain fusion, we propose a Neighborhood-Hypergraph Attention encoder, termed NAHA, which integrates Neighborhood Attention to handle local dependencies and Hypergraph Attention to capture global relationships. The encoder takes image as input $\bm{X}\in \mathbb{R} ^{H\times W\times 3}$. Here, $H$ and $W$ representthe height and width of the image. As shown in Fig.~\ref{fignaha}, the input first passes through a stem block, which partitions the input into patches and generates feature map. Then, this feature is processed by a series of Neighborhood Attention (NA) blocks and downsampling blocks to generate multi-level features $\{\bm{F}^\prime_i \in \mathbb{R}^{\frac{H}{2^{i+1}} \times \frac{W}{2^{i+1}} \times C_i}\}_{i=1}^{4}$. Then, multi-level features $\{\bm{F}^\prime_i\}_{i=2}^{4}$ are fed into a Hypergraph Attention (HA) block for global hypergraph enhancement, which captures global many-to-many relationships, producing enhanced features $\{\hat{\bm{F}}^\prime_i \in \mathbb{R}^{\frac{H}{2^{i+1}} \times \frac{W}{2^{i+1}} \times C_i}\}_{i=2}^{4}$. Finally, the enhanced features $\{\hat{\bm{F}}^\prime_i\}_{i=2}^{4}$ are added back to $\{\bm{F}^\prime_i\}_{i=2}^{4}$ to obtain the final multi-level features $\{\bm{F}_i\}^4_i=2$. The process can be formulated as follows:
\begin{equation}
    \begin{gathered} 
        \bm{F}^\prime_1=\text{NA}_1(\operatorname{stem}(\bm{X})), \quad \bm{F}^\prime_i=\text{NA}_i(\operatorname{Down}(\bm{F}^\prime_{i-1})) \\ 
        (\hat{\bm{F}}^\prime _2,\hat{\bm{F}}^\prime _3,\hat{\bm{F}}^\prime _4) =\text{HA}(\bm{F}^\prime_2, \bm{F}^\prime_3, \bm{F}^\prime_4)  \quad i\in\{2,3,4\}\\
        \bm{F}_1=\bm{F}^\prime_1, \quad \bm{F}_{i}=\bm{F}^\prime_i + \hat{\bm{F}}^\prime _i\\
    \end{gathered}
\end{equation}
Where $\text{NA}(\cdot)$ and $\text{HA}(\cdot)$ denote the Neighborhood Attention and Hypergraph Attention calculations, respectively.

\begin{figure}
    \centering
    \includegraphics[width=\columnwidth]{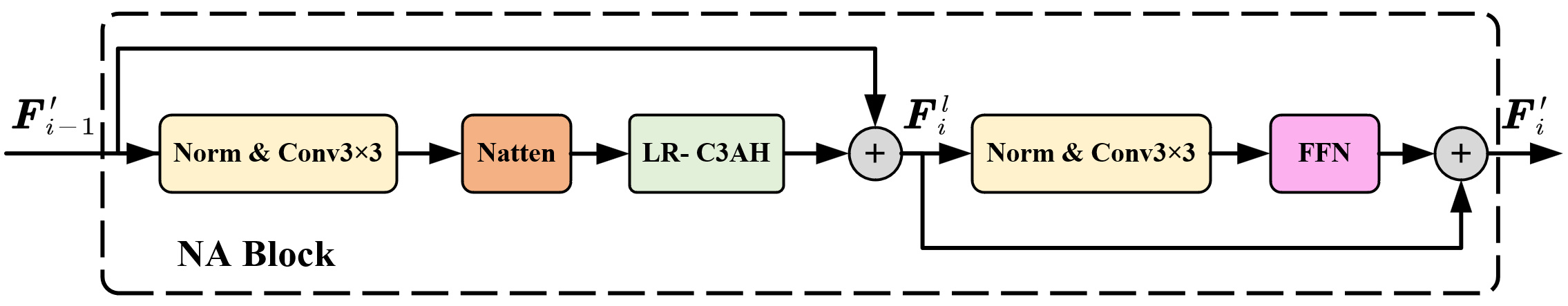}
    \caption{The NA block in the NAHA encoder.}
    \label{figna}
\end{figure}

\textbf{Neighborhood Attention Block (NA)}. As shown in Fig.~\ref{figna}, the design of the NA block is inspired by SegMAN \cite{fu2025segman}. In contrast to SegMAN, which relies on SS2D \cite{liu2024vmamba} for global modeling, our method adopts LR-C3AH \cite{yang2026m2i2ha} as an alternative and restricts its application to the last block alone. Such a strategy effectively lowers computational cost while simultaneously improving the capacity for global context modeling (details are provided in Section \ref{sec:ablation_naha}). Specifically, in the NA block, the input $\bm{F}^\prime_{i-1}$ is first normalized and passed through a $3\times 3$ convolution. Then, NATTEN \cite{hassani2023neighborhood} and LR-C3AH blocks are employed to model local relationships and global semantics, respectively. The output is then added back to $\bm{F}^\prime_{i-1}$ via a residual connection to produce $\bm{F}^l_i$. Subsequently, $\bm{F}^l_i$ is normalized, processed by another $3\times 3$ convolution, and then passed through a simple depthwise convolutional feed-forward network (FFN) \cite{xie2021segformer} to perform local spatial feature enhancement, yielding the final output $\bm{F}^\prime_{i}$.

\textbf{Hypergraph Attention Block (HA)}. The HA block follows previous work \cite{yang2026m2i2ha}. Here, it is integrated into the encoder for multi-scale feature aggregation and global relationship modeling. In the HA block, the multi-scale features $\{\bm{F}^\prime_i\}^4_{i=2}$ are aggregated to generate high-order global information, which then undergoes a size transformation to yield the enhanced features $\{\hat{\bm{F}}^\prime_i\}^4_{i=2}$ corresponding to $\{\bm{F}^\prime_i\}^4_{i=2}$. HA block leverages low-rank decomposition and sparsification to further reduce the number of parameters and inference latency while preserving high-order modeling capability and near-linear complexity. As this is not a core contribution of this paper, we do not elaborate on the details here. Interested readers are referred to \cite{yang2026m2i2ha} for further details. 

The configurations of NAHA Encoders are presented in Table~\ref{table1}.
\begin{table}[!t]
	\caption{Configurations of the four NAHA variants. FLOPs were measured at the $224\times 224$ resolution.}
    \centering
    \resizebox{\columnwidth}{!}{   
        \begin{tabular}[htbp]{@{}c|cccc@{}}
            \hline
            NAHA Encoders   & Channels                & Depths            & Params (M)    & FLOPs (G)    \\
            \hline
            Tiny            & $[32, 64, 144, 192]$    & $[2, 2, 4, 2]$    & 3.08          & 0.78         \\
            Small           & $[48, 96, 256, 384]$    & $[2, 2, 10, 4]$   & 20.40         & 3.38         \\
            Base            & $[64, 160, 320, 512]$   & $[4, 4, 18, 4]$   & 43.22         & 8.61         \\
            Large           & $[96, 192, 432, 640]$   & $[4, 4, 24, 4]$   & 80.87         & 16.80        \\
            \hline
        \end{tabular}
    }
	\label{table1}
\end{table}

\subsection{Wavelet Transform and Enhancement}
In complex scenes, intricate textures and rapid local appearance changes make precise boundary delineation challenging. Repeated spatial aggregation and downsampling can attenuate fine structural details, leading to overly smooth predictions and inaccurate boundaries. Frequency cues provide complementary information: low-frequency components capture coarse structures, whereas high-frequency components represent sharp transitions and fine details, including boundaries and textures. However, high-frequency responses may also contain noise, highlighting the need for selective enhancement. Moreover, relying solely on frequency representations can complicate the modeling of spatially localized features, whereas processing spatial and frequency features in separate branches may introduce additional computational overhead and require explicit cross-domain alignment.

\begin{figure*}
    \centering
    \includegraphics[width=0.8\textwidth]{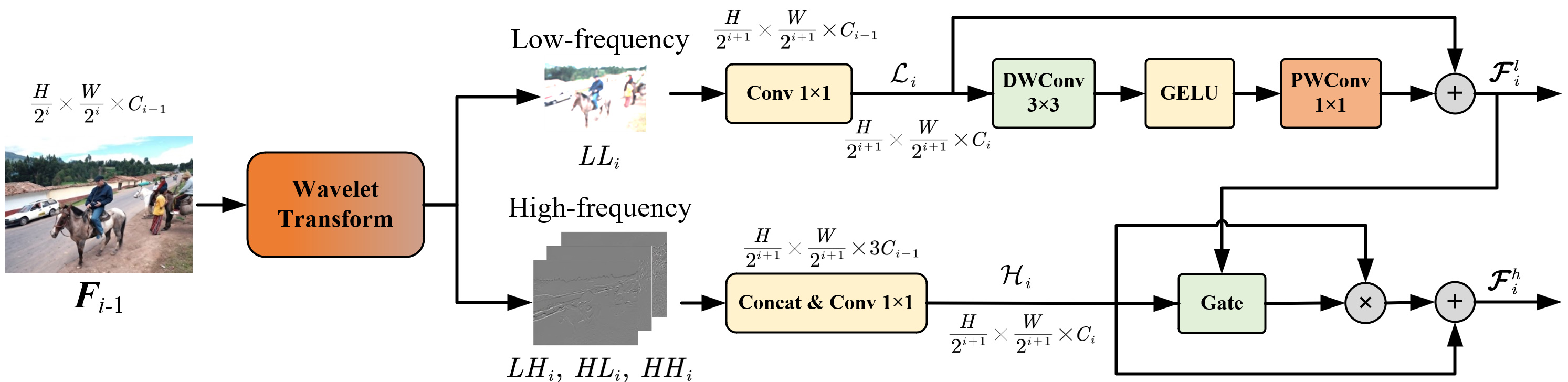}
    \caption{The Wavelet Transform and Enhancement Module.}
    \label{figwte}
\end{figure*}

We therefore introduce the Wavelet Transform and Enhancement (WTE) module to extract and enhance frequency cues directly from existing encoder features. As shown in Fig.~\ref{figwte}, WTE applies a Haar wavelet transform to the preceding-stage features $\{\bm{F}_{i-1} \in \mathbb{R}^{\frac{H}{2^{i}} \times \frac{W}{2^{i}} \times C_{i-1}}\}^4_{i=2}$, producing one low-frequency subband $LL_i \in \mathbb{R}^{\frac{H}{2^{i+1}} \times \frac{W}{2^{i+1}} \times C_{i-1}}$ and three directional high-frequency subbands $\{LH_i,HL_i, \allowbreak HH_i\}$, each subband belongs to $\mathbb{R}^{\frac{H}{2^{i+1}} \times \frac{W}{2^{i+1}} \times C_{i-1}}$. After a channel transformation, $LL_i$ is converted into $\mathcal{L}_i \in \mathbb{R}^{\frac{H}{2^{i+1}} \times \frac{W}{2^{i+1}} \times C_{i}}$. The set $\{LH_i,HL_i,HH_i\}$ is concatenated along the channel dimension and then subjected to a channel transformation to obtain $\mathcal{H}_i \in \mathbb{R}^{\frac{H}{2^{i+1}} \times \frac{W}{2^{i+1}} \times C_{i}}$. 

Then, WTE combines residual low-frequency refinement with low-frequency-conditioned high-frequency modulation to get enhanced frequency features 
\begin{equation}
    \mathcal F_i^l=\mathcal L_i+\mathcal R_l(\mathcal L_i),\qquad \mathcal F_i^h=\mathcal H_i+G_i\odot \mathcal H_i
\end{equation}
Where $\mathcal R_l$ consists of a $3\times3$ depthwise convolution, GELU, and a $1\times1$ pointwise convolution. 

Compared with independently enhancing high- and low-frequency components, we further leverage low-frequency context to modulate high-frequency responses. The motivation is that high-frequency responses can characterize local detail variations, but their response magnitude alone is insufficient to determine the importance of these variations for segmentation, since object boundaries, complex textures, and noise may all produce strong high-frequency responses. Low-frequency features, in contrast, provide coarse-scale structure and regional context, offering complementary evidence for judging the relevance of high-frequency details. Therefore, we jointly use low-frequency context and high-frequency local cues to generate spatial gating, so that the enhancement or attenuation of high-frequency responses simultaneously accounts for local variations and their surrounding structural environment, thereby achieving context-aware detail modulation. 
\begin{equation}
    \begin{gathered}
        B_i^h=\left[\operatorname{Mean}_{c}(|\mathcal H_i|),\operatorname{Max}_{c}(|\mathcal H_i|)\right] \in \mathbb{R}^{H_i \times W_i \times 2} \\
        G_i=\left(2\sigma\left(\operatorname{Conv}_{3\times3}\left[P_l(\mathcal F_i^l),B_i^h\right]\right)-1\right) \in \mathbb{R}^{H_i \times W_i \times 1}
    \end{gathered}
\end{equation}
Where $P_l$ is a $1\times1$ projection. The resulting single-channel gate is broadcast across feature channels, enabling position-dependent attenuation or amplification of high-frequency responses while maintaining spatial resolution.

The resulting frequency features subsequently guide hypergraph-based spatial feature enhancement and dual-domain fusion. This design provides a lightweight pathway for transferring frequency cues across stages and adaptively incorporating them into deeper representations, without requiring an independent frequency encoder.

\subsection{Frequency-guided Dual-domain Fusion}
\begin{figure*}
    \centering
    \includegraphics[width=0.9\textwidth]{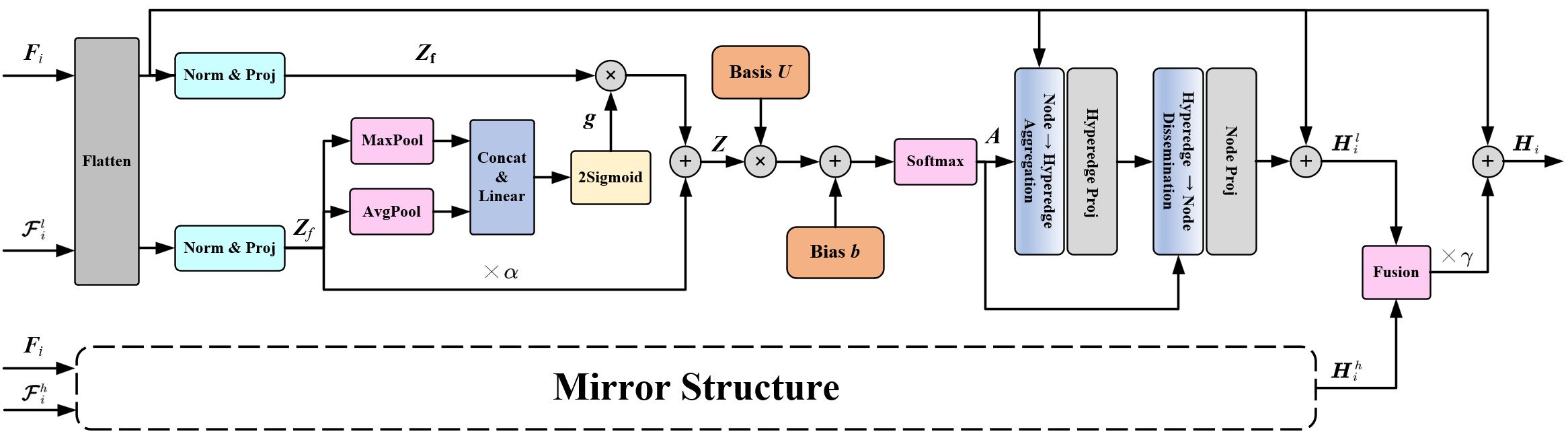}
    \caption{The Frequency-guided Dual-domain Fusion Module. Mirror Structure denotes that the high-frequency and low-frequency branches have identical architectures but do not share network parameters.}
    \label{figfgdf}
\end{figure*}

Simple addition or concatenation provides limited explicit modeling of interactions between spatial- and frequency-domain features, whereas full cross-attention between dense feature maps can introduce substantial computational overhead. To enable effective dual-domain interaction at low computational cost, we propose frequency-guided hypergraph fusion as shown in Fig.~\ref{figfgdf}. Compared with spatial-only hypergraph construction, our mechanism conditions node-hyperedge associations on both global frequency context and aligned local frequency cues. Global context adaptively modulates spatial representations, while local cues provide position-dependent guidance for node grouping. The resulting associations support spatial feature aggregation, nonlinear processing, and redistribution through a compact set of hyperedges, enabling frequency-conditioned interaction at the group level. We further employ low-rank projections to improve parameter efficiency. Each FGDF module contains parallel low- and high-frequency guided hypergraphs. Taking one branch as an example, the detailed formulation is provided below.

\textbf{Definition 1. Node Representation.} Given the flatten spatial feature matrix $\bm{F}_i=[\mathbf{f}_1,...,\mathbf{f}_N]\in \mathbb{R}^{N_i\times C_i}$ and flatten frequency feature matrix $\mathcal{F}_i=[f_1,...,f_N]\in \mathbb{R}^{N_i\times C_i}$ of stage $i$, where $N_i=H_i\times W_i$ is the number of nodes and $C_i$ is the feature dimension. For simplicity, the variable names are kept unchanged here. 

For the hyperedge prototype matrix, we introduce a low-rank node projection matrix $\bm{V}^1_i\in \mathbb{R}^{r\times C_i}$ and frequency projection matrix $\bm{V}^2_i\in \mathbb{R}^{r\times C_i}$, a low-rank hyperedge basis matrix $\bm{U}_i\in \mathbb{R}^{M\times r}$, and an edge-wise bias vector $\bm{b}_i\in \mathbb{R}^M$, where $r\ll  C_i$ denotes the low-rank dimension. The spatial and frequency features are first normalized and then projected into the $r$-dim low-rank space
\begin{equation}
    \begin{gathered}
        \bm{Z}^\mathbf{f}_i=\operatorname{LN}_\mathbf{f}(\bm{F}_i) {\bm{V}^1_i}^\text{T} \in \mathbb{R}^{N_i \times r}\\
        \bm{Z}^f_i=\operatorname{LN}_f(\mathcal{F}_i) {\bm{V}^2_i}^\text{T} \in \mathbb{R}^{N_i \times r}
    \end{gathered}
\end{equation}
Where $\operatorname{LN}(\cdot)$ represents layer normalization.

Unlike the spatial-only hypergraph construction which uses the original spatial information for context generation, here we use frequency information for context information generation and modulate the spatial information. Specifically, global average pooling and global max pooling are applied to the low-rank frequency representations
\begin{equation}
    \bm{C}_i=\left[\operatorname{MeanPool}(\bm{Z}^f_i);\operatorname{MaxPool}(\bm{Z}^f_i)
    \right]\in\mathbb{R}^{1\times 2r}
\end{equation}

A lightweight multilayer perceptron is then used to generate a rank-wise context gate
\begin{equation}
    \bm{g}_i=2\sigma\left(\operatorname{MLP}(\bm{C}_i)\right)\in\mathbb{R}^{1\times r}
    \label{eq:rank_gate}
\end{equation}
Where $\sigma(\cdot)$ denotes the sigmoid activation function. 

Finally, the context-enhanced low-rank node representations are obtained by rank-wise modulation of spatial node information together with scaled frequency feature information. This reduces computational cost while not only fusing spatial and frequency information but also effectively preserving the positional correspondence between frequency and spatial information
\begin{equation}
    \bm{Z}_i = \bm{Z}^\mathbf{f}_i\odot \bm{g}_i + \alpha \bm{Z}^f_i
    \label{eq:rank_modulation}
\end{equation}
Where $\odot$ denotes element-wise multiplication and $\bm{g}_i$ is broadcast across all nodes. $\alpha$ is a learnable scalar parameter. Intuitively, the term $\bm{Z}^\mathbf{f}_i\odot \bm{g}_i$ represents global frequency-conditioned spatial modulation, while the term $\alpha \bm{Z}^f_i$ injects local frequency representations aligned with spatial positions, thereby achieving the fusion of the spatial and frequency domains.

\textbf{Definition 2. Participation matrix.} Based on the context-enhanced low-rank node representations $\bm{Z}_i$, the low-rank hyperedge basis matrix $\bm{U}_i$, and the edge-wise bias vector $\bm{b}_i$ defined in Definition 1, the association score matrix is defined as 
\begin{equation}
        \bm{S}_i = \frac{\bm{Z}_i\bm{U}_i^\text{T}}{\sqrt{r}} + \bm{b}_i \in \mathbb{R}^{N_i\times M}
\end{equation}
Where $\bm{S}_i$ is the raw association score matrix, representing the weight of nodes relative to edges. 

Then, participation matrix $\bm{A}_i$ can be defined as
\begin{equation}
        \bm{A}^{j,k}_i = \frac{\exp (\bm{S}^{j,k}_i)}{\sum_{m=1}^{M}\exp (\bm{S}^{j,m}_i)}
\end{equation}
Where $\bm{A}_i\in \mathbb{R}^{N_i\times M}$ is the participation matrix, representing the normalized association between nodes and hyperedges.

\textbf{Definition 3. Node $\rightarrow$ Hyperedge Aggregation.} Based on participation matrix $\bm{A}_i$ and raw spatial feature matrix $\bm{F}_i$, the aggregated feature of hyperedge is defined as
\begin{equation}
    \begin{gathered}
        \bm{W}_i=(\bm{D}^{e}_{i})^{-1}{\bm{A}_i}^{\text{T}}\bm{F}_i \\
        \bm{D}^{e}_{i} = \text{diag}(\sum_{j=1}^{N_i}\bm{A}^{j,1}_i,...,\sum_{j=1}^{N_i}\bm{A}^{j,M}_i)
    \end{gathered}
\end{equation}
Where $\bm{W}_i\in \mathbb{R}^{M\times C_i}$ is the hyperedge aggregation matrix.

\textbf{Definition 4. Hyperedge $\rightarrow$ Node Dissemination.} The updated node matrix is defined as
\begin{equation}
        \bm{H}^v_i = \bm{F}_i + f_n(\bm{A}_i f_e({\bm{W}_i})) \qquad (v=l, h)
\end{equation}
Where, $f_n(\cdot)$ and $f_e(\cdot)$ denote the learned nonlinear projection functions on nodes and hyperedges, respectively. $\bm{H}^v_i$ denotes the final enhanced feature after dual-domain fusion, where $v=l$ and $v=h$ denote the low- and high-frequency branches, respectively.

Finally, $\bm{H}^{l}_i$ and $\bm{H}^{h}_i$ are concatenated and fused. The fused feature is then added to the original spatial input via a residual connection, yielding the final output $\bm{H}_i$:
\begin{equation}
    \bm{H}_i = \bm{F}_i + \gamma\operatorname{Conv}_{3\times 3}\left(\operatorname{Concat}(\bm{H}^{l}_i, \bm{H}^{h}_i) \right) 
\end{equation}
where $\gamma$ is a learnable scalar parameter.

Unlike the conventional dual-domain decoupled fusion paradigms adopted in DDFNet \cite{li2025dual} and SFFNet \cite{yang2024sffnet}, which transform spatial- and frequency-domain features in parallel and decompose both into low- and high-frequency components, our method applies Haar decomposition to preceding-stage encoder features and uses the resulting frequency representations to guide current-stage spatial grouping and feature propagation. This design not only alleviates the misalignment caused by parallel transformations across the spatial and frequency domains but also reduces computational overhead.

\subsection{HyperDecoder and Prediction Heads}
\begin{figure*}
    \centering
    \includegraphics[width=0.9\textwidth]{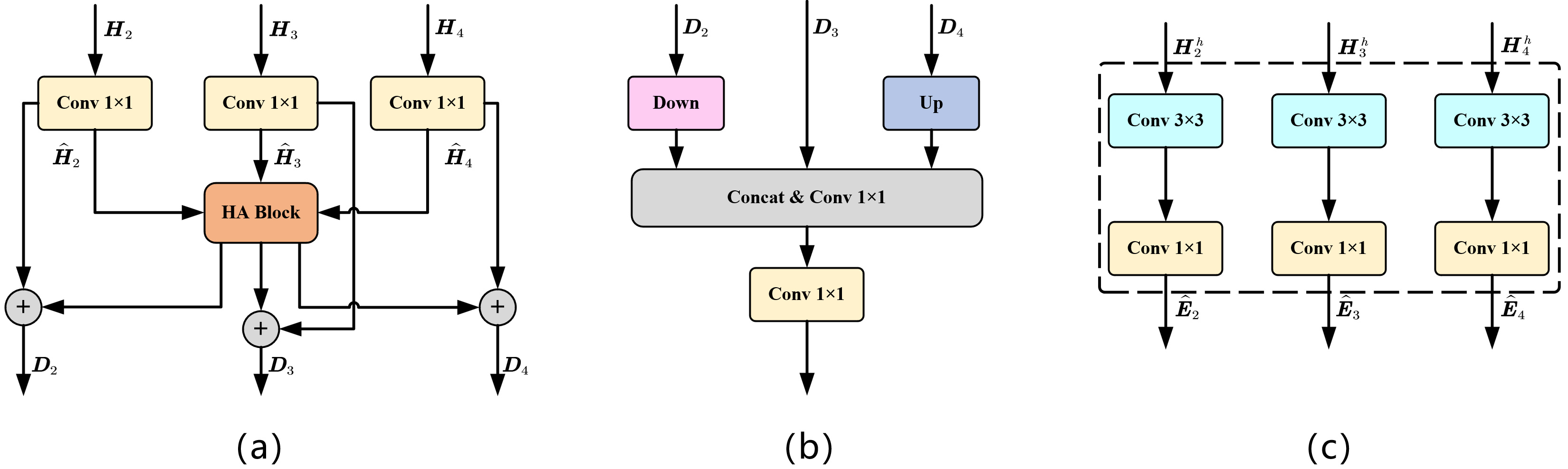}
    \caption{(a) HyperDecoder. (b) Segmentation head. (c) Multi-level edge prediction head.}
    \label{fig4}
\end{figure*}

To integrate the enhanced multi-level features, we design a HyperDecoder equipped with a hypergraph attention (HA) block. As shown in Fig.~\ref{fig4}(a), the features $\{\bm{H}_2,\bm{H}_3,\bm{H}_4\}$ are first projected to a common channel dimension, yielding $\{\hat{\bm{H}}_2, \hat{\bm{H}}_3,\hat{\bm{H}}_4\}$. The HA block models high-order relationships across these feature levels and produces multi-level enhanced representations that match the spatial resolution of each level. These refined features are then added to the corresponding $\hat{\bm{H}}_i$, yielding the refined features $\{\bm{D}_2,\bm{D}_3,\bm{D}_4\}$.

As illustrated in Fig.~\ref{fig4}(b), $\bm{D}_2$ and $\bm{D}_4$ are downsampled and upsampled, respectively, to match the spatial resolution of $\bm{D}_3$. The aligned features are concatenated along the channel dimension and fused by a $1\times1$ convolution. A subsequent $1\times1$ classifier produces pixel-wise class predictions.

In parallel, the high-frequency-guided spatial features $\{\bm{H}^h_2,\bm{H}^h_3,\bm{H}^h_4\}$ are fed into a multi-level edge prediction head, as shown in Fig.~\ref{fig4}(c), to produce boundary predictions $\{\hat{\bm{E}}_2,\hat{\bm{E}}_3,\hat{\bm{E}}_4\}$ at different scales. These predictions are supervised by an auxiliary edge loss, encouraging the network to use high-frequency cues to learn boundary-sensitive representations.   

\subsection{Loss function}
To design the loss function, we adopt a joint formulation consisting of a pixel-level loss $L_{\text{pixel}}$ and an auxiliary boundary loss $L_{\text{edge}}$. The pixel-wise cross-entropy loss supervises semantic classification, while the auxiliary boundary loss provides explicit supervision for boundary localization and encourages boundary-sensitive representations.

The pixel-wise loss is computed using the cross-entropy loss between the predicted mask and the ground truth label, defined as follows:
\begin{equation}
L_{\text{pixel}} = -\frac{1}{HW} \sum_{i=1}^{H} \sum_{j=1}^{W} \sum_{c=1}^{C} Y(i,j,c) \log(\hat{Y}(i,j,c))
\end{equation}
Where $H$ and $W$ are the height and width of the feature map, $C$ is the number of classes, $Y(i,j,c)$ is the one-hot representation of the ground truth label $Y(i,j)$, and $\hat{Y}(i,j,c)$ represents the predicted probability of class $c$ at spatial location $(i,j)$. This loss aims to enable the network to accurately classify each pixel.

For the auxiliary boundary loss, we employ deep supervision, ensuring the high-frequency information at each layer can adapt correctly. The loss for multi-level edge predictions is summed and averaged, which is defined as:
\begin{equation}
L_{\text{edge}} = \frac{1}{KHW}\sum_{k=1}^{K}\sum_{i=1}^{H}\sum_{j=1}^{W}|E(i,j) - \text{I}(\hat{E}_k)(i,j)|
\end{equation}
Where $E(i,j)$ is the binary edge ground truth label. It is generated from the semantic ground truth $Y(i,j)$ following the boundary generation method in DDFNet \cite{li2025dual}. This step is carried out offline in the data preprocessing stage, so no differentiable operation is involved and no gradient needs to be propagated through it. $\hat{E}_k(i,j)$ denotes the $k$-th binary edge prediction, and $\text{I}(\cdot)$ denotes the interpolation operation used to resize the predicted edge mask to the original $H\times W$ resolution. This loss constrains the network to effectively utilize high-frequency information, enhancing the continuity and precision of the edge predictions.

The total loss function of the network is defined as:
\begin{equation}
L_{\text{total}} = \lambda  L_{\text{pixel}} + (1-\lambda) L_{\text{edge}}
\end{equation}
Where $\lambda$ is a hyperparameter that controls the trade-off between semantic classification and boundary localization. An appropriate value of $\lambda$ is crucial for balancing the pixel-level prediction accuracy and the smoothness of the boundaries.

\section{Experiments and Results}
In this section, we conduct experiments to demonstrate the effectiveness of the proposed FGD2HF. First, we briefly introduce the dataset, experimental setup, and evaluation metrics. Next, we present detailed comparisons between our method and several benchmark methods on relevant datasets. Finally, we perform ablation studies to verify the necessity of each proposed module.

\subsection{Settings}
\textbf{Datasets.} To comprehensively evaluate the performance of the proposed FGD2HF, experiments were conducted on three widely recognized public benchmark datasets: Pascal VOC (SBD) \cite{hariharan2011semantic}, Cityscapes \cite{cordts2016cityscapes} and ADE20K \cite{zhou2017scene}. 

\textbf{Baselines.} We selected representative strong baselines for comparison, including methods without frequency information, such as DeepLabV3 \cite{chen2017rethinking}, DeepLabV3+ \cite{chen2018encoder}, SegFormer \cite{xie2021segformer}, SegNeXt \cite{guo2022segnext}, VWFormer \cite{yan2024multi}, EDAFormer \cite{yu2024embedding} and  SegMAN \cite{fu2025segman}, as well as methods that employ frequency information, such as FsaNet \cite{zhang2023fsanet}, FADC \cite{chen2024frequency}, SegAliasing \cite{chen2024semantic}, FreqFusion \cite{chen2024frequencyaware}, SFM \cite{chen2023spatialfrequency}, FDConv \cite{chen2025frequency}, and SFFNet \cite{yang2024sffnet}.

\textbf{Implementation details.} All experiments were conducted on a Linux server using PyTorch, equipped with an AMD Ryzen 9 9950X CPU and two NVIDIA RTX Pro 6000 GPUs (96GB). For the unified inference-speed benchmark reported in Table 5, all models were evaluated on a single NVIDIA A100 GPU. To ensure fair comparison, we keep the common parameters consistent across all experiments as much as possible. We applied standard data augmentation techniques, such as random flip, random crop and random scale. However, strong augmentations (e.g., Mosaic, Copy-Paste, CutMix) were prohibited. Inference was performed in FP32 mode without augmentation. Inference efficiency is evaluated using frames per second (FPS), calculated as the number of processed images divided by the total model forward-pass time. Measurements use synthetic input tensors stored on the GPU and exclude data loading, input preparation, and post-processing. The default training and inference configurations are summarized in Table~\ref{table2}.

\begin{table}[!t]
    \centering
    \caption{Default training and inference configurations.}
    \resizebox{0.8\columnwidth}{!}{
        \begin{tabular}{@{}c|c@{}}
            \toprule
            \textbf{Items} & \textbf{Values} \\
            \midrule
            \multicolumn{2}{c}{\textbf{Training Settings}} \\
            \midrule
            Image size & $512 \times 512$ ($1024 \times 1024$ for Cityscapes) \\
            Batch size & $32$  \\
            AMP & True \\
            \midrule
            \multicolumn{2}{c}{\textbf{Optimization}} \\
            \midrule
            Optimizer & AdamW \\
            Initial LR & $0.0006$ \\
            Backbone LR scale & $0.1$ \\
            Weight decay & $0.01$ \\
            beta1 & $0.9$ \\
            beta2 & $0.999$ \\
            \midrule
            \multicolumn{2}{c}{\textbf{Scheduler}} \\
            \midrule
            Scheduler & PolyLR \\
            Power & $0.9$ \\
            \midrule
            \multicolumn{2}{c}{\textbf{Data Augmentation}} \\
            \midrule
            HSV-H & $0.01$ \\
            HSV-S & $0.4$ \\
            HSV-V & $0.25$ \\
            Translate & $0.05$ \\
            \midrule
            \multicolumn{2}{c}{\textbf{Loss Weights}} \\
            \midrule
            CE loss weights (pixel) & $0.8$ \\
            Geo loss (edge) & $0.2$ \\
            \bottomrule
        \end{tabular}%
    }
    \label{table2}
\end{table}

\textbf{Metrics.} We adopt mean Intersection over Union (mIoU) as the primary metric for evaluating overall segmentation performance. We further assess boundary quality using mean Boundary IoU (mBIoU) and mean boundary F1 score (mBF1). For mBIoU, the boundary-band width is set to 2\% of the image diagonal, while mBF1 uses one-pixel-wide boundaries with a matching tolerance of 0.8\% of the image diagonal. Boundaries are extracted separately for each semantic class, and pixels with the ignore label are excluded from evaluation. For both metrics, boundary statistics are accumulated across the entire evaluation set for each class before computing the class-wise scores, which are then averaged over valid classes.

\subsection{Performance of NAHA Encoder}
To evaluate the feature extraction capability of our proposed NAHA encoder, we pretrain it on the ImageNet-1k dataset for image classification. The classification results are compared with the backbones used in comparative segmentation methods (e.g., MobileNetV3 \cite{howard2019searching}, MIT \cite{xie2021segformer}, MSCAN \cite{guo2022segnext}, EFT \cite{yu2024embedding}, SagMAN-Encoder \cite{fu2025segman}) as well as other visual representation backbones (e.g., Swin \cite{liu2021swin}, ConvNeXt \cite{liu2022convnet}, Vim \cite{zhu2024vision}, VMamba \cite{liu2024vmamba}, EfficientVMamba \cite{pei2025efficientvmamba}, SparX-Mamba \cite{lou2025sparx}). The comparison results are shown in Table~\ref{table3}.

\begin{table}[!t]
    \caption{Comparison of NAHA encoder with state-of-the-art backbones on the ImageNet-1K datasets.}
    \centering
    \resizebox{0.9\columnwidth}{!}{
        \begin{tabular}{@{}cccc@{}}
            \hline
            Models                   & Params (M) & GFLOPs & Top-1 Acc (\%)                         \\
            \hline
            MIT-B0                   & 3.8        & 0.6    & 70.4                                   \\
            MSCAN-T                  & 4.2        & 0.9    & 75.7                                   \\
            EFT-T                    & 3.7        & 0.6    & 72.5                                   \\
            MobileNetV3-L            & 5.2        & 0.5    & 75.2                                   \\
            SegMAN-Encoder-T         & 3.5        & 0.7    & 76.2                                   \\
            EfficientVMamba-T        & 6.0        & 0.8    & \textcolor{green}{76.4}                \\
            NAHA-T (Ours)            & 3.1        & 0.8    & \textcolor{red}{77.1}                  \\
            \hline
            EfficientVMamba-S        & 11.0       & 1.3    & 78.7                                   \\
            \hline
            MIT-B2                   & 24.0       & 4.0    & 81.6                                   \\
            MSCAN-B                  & 27.0       & 4.4    & 83.0                                   \\
            EFT-B                    & 26.0       & 4.2    & 82.3                                   \\
            SegMAN-Encoder-S         & 26.0       & 4.1    & \textcolor{red}{83.6}                  \\
            Swin-T                   & 28.0       & 4.5    & 81.1                                   \\
            ConvNeXt-T               & 28.0       & 4.5    & 82.1                                   \\
            Vim-S                    & 26.0       & -      & 81.5                                   \\
            VMamba-T                 & 29.0       & 4.9    & 82.6                                   \\
            SparX-Mamba-T            & 27.0       & 5.2    & \textcolor{green}{83.5}                \\
            EfficientVMamba-B        & 33.0       & 4.0    & 81.8                                   \\
            NAHA-S (Ours)            & 20.4       & 3.4    & 83.2                                   \\ 
            \hline
            MIT-B3                   & 45.0       & 6.9    & 83.1                                   \\
            MSCAN-L                  & 45.0       & 9.1    & 83.9                                   \\
            SegMAN-Encoder-B         & 45.0       & 9.9    & \textcolor{green}{84.4}                \\
            Swin-S                   & 50.0       & 8.7    & 83.2                                   \\
            ConvNeXt-S               & 50.0       & 8.7    & 83.1                                   \\
            VMamba-S                 & 50.0       & 8.7    & 83.6                                   \\
            SparX-Mamba-S            & 47.0       & 9.3    & 84.2                                   \\
            NAHA-B (Ours)            & 43.2       & 8.6    & \textcolor{red}{84.5}                  \\
            \hline
            MIT-B5                   & 82.0       & 11.8   & 83.8                                   \\
            SegMAN-Encoder-L         & 81.0       & 16.8   & \textcolor{green}{85.3}                \\
            Swin-B                   & 88.0       & 15.4   & 83.5                                   \\
            ConvNeXt-B               & 89.0       & 15.4   & 83.8                                  \\
            Vim-B                    & 98.0       & -      & 83.2                                   \\
            VMamba-B                 & 89.0       & 15.4   & 83.8                                   \\
            SparX-Mamba-B            & 84.0       & 15.9   & 84.5                                   \\
            NAHA-L (Ours)            & 80.9       & 16.8   & \textcolor{red}{85.4}                  \\
            \hline
        \end{tabular}
    }
    \label{table3}
\end{table}

The results demonstrate that the NAHA encoder achieves competitive performance compared with state-of-the-art backbones, offering a favorable balance between model complexity and accuracy. In the comparison with lightweight models, NAHA-T achieves the best classification accuracy of 77.1\% with the lowest parameter count of 3.1 M and a comparable computational cost of 0.8 GFLOPs, outperforming the recently proposed Mamba-based EfficientVMamba-T by 0.7\%. NAHA-S achieves an accuracy of 83.2\%. Although this is not the highest accuracy, it uses the fewest parameters and FLOPs (21\% and 17\% fewer than the best SegMAN-S model, respectively). This is because, in our design, we aimed to increase the network width as uniformly as possible, thereby achieving a better balance between accuracy and computational cost. In addition, NAHA-B and NAHA-L also deliver highly competitive classification accuracies of 84.5\% and 85.4\%, respectively. These results indicate the efficiency and effectiveness of NAHA in feature extraction.

\subsection{Comparison with State-of-The-Art Methods}
This section presents a comprehensive comparison between our proposed FGD2HF and state-of-the-art (SOTA) segmentation methods across three widely adopted datasets. It should be noted that all of our reproduced results are in strict accordance with the configurations of the original papers, and for any cases that are not readily reproducible, we use the data as originally reported. Throughout the training process, we kept the hyperparameter settings largely consistent. Specifically, the training resolution was set to $512\times512$ for the SBD and ADE20K datasets, and $1024\times1024$ for Cityscapes. For validation and inference, we adopted $512\times512$ for SBD and ADE20K, and $2048\times1024$ for Cityscapes. The model's GFLOPs are reported based on the inference resolutions. Furthermore, as different codebases employ distinct data loading and training paradigms, we standardized the training duration according to their respective implementations: for iteration-based codebases, we followed previous literature and set the duration to 160k iterations; for epoch-based codebases, we uniformly set the training length to 100 epochs.

\textbf{Segmentation Result.} Table~\ref{table4} summarizes the performance of FGD2HF against several leading methods, including both traditional spatial-domain approaches and recent frequency-aware models. Among lightweight models, FGD2HF-T achieves the best performance, with mIoU scores of 79.8\% on the SBD dataset, 79.7\% on Cityscapes, and 40.5\% on ADE20K. Compared with frequency-based methods, FGD2HF-T improves upon FreqFusion-SegNext-T by approximately 0.5\% on Cityscapes with 11.8\% fewer GFLOPs. Among non-frequency-based methods, FGD2HF-T also outperforms the recently proposed SegMAN-T by 0.8\% on Cityscapes, while maintaining higher computational efficiency. FGD2HF-S further achieves the competitive performance among models of smaller scale, obtaining mIoU scores of 83.6\% on the SBD dataset, 82.0\% on Cityscapes, and 48.1\% on ADE20K. On Cityscapes, FGD2HF-S improves over FreqFusion-SegFormer-B1 by approximately 1.3\%, while using 9.5\% fewer GFLOPs than frequency-based methods. Among non-frequency-based methods, FGD2HF-S surpasses the recently proposed SegMAN-S by 0.2\%, while still preserving a better balance between segmentation accuracy and computational cost. FGD2HF-B achieves the second-best performance on SBD with an mIoU of 84.0\%, only 0.2 percentage points lower than SegMAN-B. However, it obtains the best results on Cityscapes and ADE20K, reaching 83.2\% and 48.7\%, respectively. FGD2HF-L achieves the best performance on the three datasets, with scores of 84.7\%, 83.8\%, and 50.3\%, respectively.

\textbf{Boundary Prediction Result.} Table~\ref{table4} further demonstrates the competitive boundary segmentation performance of FGD2HF across different model scales. Among lightweight models, FGD2HF-T achieves the highest mBIoU and mean boundary F1 score (mBF1) on all three datasets, with scores of 71.3\%/71.6\% on SBD, 74.9\%/85.4\% on Cityscapes, and 29.5\%/38.0\% on ADE20K. FGD2HF-S improves mBIoU and mBF1 over SegMAN-S by 0.7 and 0.5 percentage points on SBD, 0.4 and 0.8 points on Cityscapes, and 1.2 and 0.9 points on ADE20K, respectively, while requiring fewer GFLOPs. FGD2HF-B leads its comparison group on both boundary metrics across all three datasets, achieving 78.3\%/77.8\% on SBD, 79.7\%/90.0\% on Cityscapes, and 39.2\%/49.4\% on ADE20K. On SBD, it surpasses SegMAN-B on both boundary metrics despite having a slightly lower mIoU, highlighting the complementary information provided by boundary evaluation. FGD2HF-L also outperforms SegMAN-L on both boundary metrics across all three datasets and achieves the highest mBIoU and mBF1 among the compared methods on ADE20K, reaching 40.3\% and 51.5\%, respectively.

The results also show that incorporating frequency information does not necessarily yield better boundary segmentation than spatial-only approaches. For example, SFFNet performs below SegMAN-S on both boundary metrics on ADE20K. Since these methods differ in their encoders and fusion strategies, such comparisons reflect the performance of the complete architectures rather than the isolated contribution of frequency processing. Boundary quality depends on both the encoder representations and how frequency cues are incorporated into subsequent feature processing. This consideration motivates our design of the NAHA encoder, which combines neighborhood and hypergraph attention to capture local structural details and global semantic context, providing informative representations for frequency extraction and dual-domain fusion.

\begin{table*}[!t]
    \caption{Comparison of FGD2HF with state-of-the-art methods on the Pascal VOC (SBD), Cityscapes, and ADE20K datasets. $^{\dagger}$ indicates the results reported in the original papers. $-$  indicates that the value is not reported in the original paper or was not measured.}
    \centering
    \resizebox{\textwidth}{!}{   
        \begin{tabular}{@{}c|c|c|cccc|cccc|cccc@{}}
            \hline
            \hline
            \multirow{2}{*}{Method} & \multirow{2}{*}{Cat.} & \multirow{2}{*}{Params (M)} & \multicolumn{4}{c|}{SBD} & \multicolumn{4}{c|}{Cityscapes} & \multicolumn{4}{c}{ADE20K} \\
            & & & GFLOPs & mIoU & mBIoU & mBF1 & GFLOPs & mIoU & mBIoU & mBF1 & GFLOPs & mIoU & mBIoU & mBF1 \\
            \hline
            \hline
            DeepLabV3-MobileNet (CVPR 2017) \cite{chen2017rethinking}   & \multirow{6}{*}{Non-Fre.} & 5.1  & 6.1   & 70.1  &58.2&57.1& 47.2   & 68.8 &  62.5  &  76.1  & 6.1   & 34.7 &  24.4  &  31.2  \\
            DeepLabV3+-MobileNet (ECCV 2018) \cite{chen2018encoder}     &                           & 5.2  & 17.6  & 71.0  &59.0&57.4& 135.1  & 72.1 &63.2&76.3& 17.6  & 35.0 &  25.2  &  31.9  \\
            SegFormer-B0 (NeurIPS 2021) \cite{xie2021segformer}         &                           & 3.8  & 8.4   & 68.3  &54.5&52.6& 121.1  & 72.4 &65.6&76.7& 8.4   & 36.4 &  25.8  &  32.4  \\
            SegNeXt-T (NeurIPS 2022) \cite{guo2022segnext}              &                           & 4.3  & 6.6   & 77.9  &65.7&64.7& 56.0   & 79.0 &74.0&83.9& 6.6   & 38.0 & 27.0 & 34.5 \\
            EDAFormer-T (ECCV 2024) \cite{yu2024embedding}              &                           & 5.0  & 5.6   & 75.9  &64.9&63.3& 151.5  & 76.9 &71.5&82.3& 5.6   & 38.1 & 26.9 &  34.6  \\
            CGRSeg-T (ECCV 2024) \cite{ni2024context}                   &                           & 9.4  & 4.8   & 78.8  &70.0&69.7& 65.5   & 78.3 &73.1&83.6& 4.8   & 38.7 &27.9& 34.3 \\
            SegMAN-T (CVPR 2025) \cite{fu2025segman}                    &        & 6.4  & 6.0   & \textcolor{green}{79.0}  &\textcolor{green}{70.2} &\textcolor{green}{70.4}& 52.5   & 78.9 &73.8&84.4& 6.0   & 38.3 &27.8&35.8\\
            \hline
            FreqFusion-SegNext-T (TPAMI 2024) \cite{chen2024frequencyaware}  & \multirow{2}{*}{Fre.} & 4.4 & - & - & -&  - &    52.4 & \textcolor{green}{79.2} &\textcolor{green}{74.5}&\textcolor{green}{84.8}& 7.0  & \textcolor{green}{40.2} & \textcolor{green}{29.3} & \textcolor{green}{37.5}\\
            FGD2HF-T (Ours)                 &      & 5.0 &  5.8  & \textcolor{red}{79.8} & \textcolor{red}{71.3} & \textcolor{red}{71.6} &  46.2 & \textcolor{red}{79.7}  & \textcolor{red}{74.9} & \textcolor{red}{85.4} & 5.8 & \textcolor{red}{40.5} &\textcolor{red}{29.5}& \textcolor{red}{38.0}\\
            \hline
            \hline
            SegFormer-B2 (NeurIPS 2021) \cite{xie2021segformer}         & \multirow{4}{*}{Non-Fre.} & 27.4  & 60.7  & 78.0 &67.4&64.7& 703.6 &  78.5 &73.5&83.9& 60.7  & 45.4 &34.0&42.4\\
            SegNeXt-B (NeurIPS 2022) \cite{guo2022segnext}              &                           & 27.6  & 34.9  & 82.0 &71.2&69.6& 275.6 &  81.7 &\textcolor{green}{77.7}&\textcolor{green}{88.2}& 34.9  & 45.6 & 35.4 & 45.1\\
            EDAFormer-B (ECCV 2024) \cite{yu2024embedding}              &                           & 29.5  & 31.9  & 81.7 &72.4&70.0& 605.4 &  80.3 &75.6&85.7& 31.9  & 47.5 & 36.5 & 46.0 \\
            CGRSeg-L (ECCV 2024) \cite{ni2024context}                   &                           & 35.7  & 16.5  & 83.4 &76.9&76.1& 199.7 &  80.2 & 75.2 & 85.1 & 16.5  & 47.3 & 36.2 & 46.1 \\
            SegMAN-S (CVPR 2025) \cite{fu2025segman}                    &                           & 30.0  & 25.1  & \textcolor{red}{83.9} & \textcolor{green}{77.2} & \textcolor{green}{76.5} & 218.4 &  \textcolor{green}{81.8} &77.6 &  88.0  & 25.1  & \textcolor{green}{47.9} &  \textcolor{green}{37.4}  &  \textcolor{green}{47.0}  \\
            \hline
            SegAliasing (ICLR 2024) \cite{chen2024semantic} $\dagger$   & \multirow{4}{*}{Fre.}     & 31.5  & 37.4  & 76.1 & -  & -  & 298.7 &  79.7 &  - &  - & 37.4  & 40.4  &  -  &  - \\
            FreqFusion-SegFormer-B1 (TPAMI 2024) \cite{chen2024frequencyaware} &                    & 14.1  &   -   &   -  &  - &  - & 174.3 &  80.7 & 76.9&87.3 & 18.3  & 43.9  &  33.7  & 42.0  \\
            SFFNet (TGRS 2024) \cite{yang2024sffnet}                    &                           & 34.2  & 52.3  & 79.9 & 75.4&74.8 & 421.9 &  80.2 & 76.5&87.0 & 52.3  & 44.8  &  35.5  & 45.3  \\
            FGD2HF-S (Ours)                                             &                           & 23.4  & 19.7  & \textcolor{green}{83.6} &\textcolor{red}{77.9}&\textcolor{red}{77.0} & 157.8 &   \textcolor{red}{82.0}  &\textcolor{red}{78.0}&\textcolor{red}{88.8} & 19.7  &  \textcolor{red}{48.1}   &  \textcolor{red}{38.6}  & \textcolor{red}{47.9}    \\
            \hline
            \hline
            DeepLabV3-Resnet50 (CVPR 2017) \cite{chen2017rethinking}    & \multirow{8}{*}{Non-Fre.} & 39.7  & 51.4  & 76.9 &67.1& 63.7 & 389.4 & 74.1 & 69.0& 80.1 & 51.4  & 43.6 &32.9&43.1 \\
            DeepLabV3+-Resnet50 (ECCV 2018) \cite{chen2018encoder}      &                           & 39.8  & 62.7  & 77.2 & 67.8 & 64.5& 478.8 & 75.0 & 69.4 & 80.8 & 62.7  & 44.0 &33.7&44.0 \\
            Segformer-B3 (NeurIPS 2021) \cite{xie2021segformer}         &                           & 47.2  & 77.4  & 80.4 &70.2&66.8& 949.5 & 80.1 &75.2&85.5 & 77.4  & 47.2 & 36.3 & 45.3\\
            MaskFormer (NeurIPS 2021) \cite{cheng2021per}               &                           & 42.0  & 55.0  & 81.1 &72.0&69.3 &  -    &  -   & - &  - & 55.0  & 46.5 &35.5 &46.2\\
            Mask2Former (CVPR 2022) \cite{cheng2022masked}              &                           & 47.0  & 74.2  & 82.3 &73.3&71.1 &  -    &  -   & -  & -  & 74.2  & 47.3 &37.9 &48.4 \\
            SegNeXt-L (NeurIPS 2022) \cite{guo2022segnext}              &                           & 48.8  & 70.0  & 82.1 &73.0&70.8& 578.0 & \textcolor{green}{82.6} &78.8&88.7& 70.0  & 47.8 & 37.8 & 48.2 \\
            VWFormer-T (ICLR 2024) \cite{yan2024multi}                  &                           & 44.8  & 60.0  & 82.9 &74.2 &74.0  &  -    &  -   & - &  -  & 60.0  & 47.4 &37.2  &47.9 \\
            SegMAN-B (CVPR 2025) \cite{fu2025segman}                    &                           & 51.8  & 57.8  & \textcolor{red}{84.2} &  \textcolor{green}{76.6}  & \textcolor{green}{76.7} & 479.0 & \textcolor{red}{83.2} & \textcolor{green}{79.3} & \textcolor{green}{89.3} & 57.8  &\textcolor{green}{48.4} & \textcolor{green}{38.5}   & \textcolor{green}{48.5}   \\
            \hline
            SFM-ResNet50 (TPAMI 2025) \cite{chen2023spatialfrequency} $\dagger$  & \multirow{2}{*}{Fre.}    & 50.4  &   -   &   -  & -  & -  & 221.5 & 76.0 &  -  &  -  &  -    &  -  & -  & - \\
            FGD2HF-B (Ours)                                             &                           & 47.2  & 47.2  & \textcolor{green}{84.0} & \textcolor{red}{78.3} & \textcolor{red}{77.8} & 377.4 &  \textcolor{red}{83.2}   & \textcolor{red}{79.7}   &  \textcolor{red}{90.0}   & 47.2  &   \textcolor{red}{48.7}  &  \textcolor{red}{39.2}  & \textcolor{red}{49.4}    \\
            \hline
            \hline
            DeepLabV3-Resnet101 (CVPR 2017) \cite{chen2017rethinking}   & \multirow{5}{*}{Non-Fre.} & 58.6  & 72.1  & 77.3 & 68.8&64.6 & 545.4  & 74.8 &69.5 &80.9 & 72.1  & 44.8 &33.3  & 43.7 \\
            DeepLabV3+-Resnet101 (ECCV 2018) \cite{chen2018encoder}     &                           & 58.8  & 83.4  & 78.3 & 69.2& 66.1   & 634.8  & 76.4 & 70.6 &  82.1  & 83.4  & 45.5 & 34.1 & 45.2 \\
            Segformer-B5 (NeurIPS 2021) \cite{xie2021segformer}         &                           & 84.6  & 110.0 & 81.0 &71.6&68.3& 1446.8 & 81.0 &76.1&85.9& 110.0 & 48.7 &38.2& 47.5\\
            VWFormer-S (ICLR 2024) \cite{yan2024multi}                  &                           & 66.1  & 83.7  & 83.3 & 75.5   & 74.4   &   -    &   -  & -  & -  & 83.7  & 49.6 & 38.9 & 49.0   \\
            SegMAN-L (CVPR 2025) \cite{fu2025segman}                    &                           & 92.3  & 97.2  & \textcolor{green}{84.3} &  \textcolor{green}{77.3}  & \textcolor{green}{77.3} & 796.0  & \textcolor{green}{83.3} &  \textcolor{green}{79.3} & \textcolor{green}{89.4} & 97.2  & \textcolor{green}{49.7} & \textcolor{green}{40.0} & \textcolor{green}{51.1} \\
            \hline
            FsaNet-Lin-R2 (TIP2023) \cite{zhang2023fsanet} $\dagger$    & \multirow{6}{*}{Fre.}     & 69.0  & 276.6 & 78.7 &  -  &  -  & 2212.0 & 80.6 &  -  &   - & 276.6 & 44.1 & -   &  - \\
            FsaNet-Dot-R2 (TIP2023) \cite{zhang2023fsanet} $\dagger$    &                           & 69.1  & 276.8 & 78.7 &  -  &   - & 2214.1 & 80.4 &  -  &   - & 276.8 & 43.5 &  -  &  - \\
            FADC (CVPR 2024) \cite{chen2024frequency}                   &                           & 67.4  & 230.7 & 77.4 &69.0&71.1& 1845.0 & 76.2 &  72.3  & 83.6 & 230.7 & 44.4 & 32.5 & 43.1 \\
            SFM-ResNet101 (TPAMI 2025) \cite{chen2023spatialfrequency} $\dagger$  &                 & 69.4  &   -   &   -  &  -  &  -  & 376.8  & 76.6 &  -  &  -  &  -    &  -  & -   &  -  \\
            FDConv-ResNet50 (CVPR 2025) \cite{chen2025frequency} $\dagger$  &                       & 70.0  &   -   &   -  &  -  &  -  & -      & 80.4 &  -  &  -  & 949.0 & 44.9 &  -  &  - \\
            FGD2HF-L (Ours)                                             &                           & 90.8  & 93.5  & \textcolor{red}{84.7} & \textcolor{red}{78.7}  & \textcolor{red}{78.0}   & 747.8 &   \textcolor{red}{83.8}  &  \textcolor{red}{79.5}  & \textcolor{red}{89.6}    & 93.5  &   \textcolor{red}{50.3} &  \textcolor{red}{40.3}  & \textcolor{red}{51.5}    \\
            \hline
            \hline
        \end{tabular}
    } 
    \label{table4}
\end{table*}

\subsection{Analysis of Inference Speed and GPU Memory Usage}
To evaluate model inference speed and GPU memory usage, we benchmark FGD2HF on a single NVIDIA A100 GPU using randomly generated RGB input tensors of size $512 \times 512$, with a batch size of 1 and FP32 precision. The model is set to evaluation mode, with gradient computation disabled. After 10 warm-up iterations, we record the total execution time over 1,000 forward passes and calculate FPS as the number of processed images divided by the elapsed time. Only model forward execution is timed, with input tensors already stored on the GPU. Data loading, input preparation, and post-processing are excluded to avoid discrepancies arising from different data pipelines and post-processing implementations across methods. GPU Memory denotes the peak GPU memory usage during inference.

As reported in Table~\ref{table5},  FGD2HF-T, -S, -B, and -L achieve inference speeds of 96.0, 71.7, 46.4, and 36.3 FPS, respectively, while maintaining peak GPU memory usage of 188.6, 299.7, 386.8, and 637.4 MB. Together with their segmentation accuracy on SBD, these results demonstrate a favorable trade-off between accuracy and inference efficiency.
\begin{table}[!t]
    \caption{Comparison of model complexity and inference speed on SBD.}
    \centering
    \resizebox{0.9\columnwidth}{!}{
        \begin{tabular}{@{}c|ccccc@{}}
            \hline
            Models                            & Params (M) & GFLOPs & mIoU (\%)  &   FPS (Hz) &   GPU Memory (MB) \\
            \hline
            Segformer-B0                      & 3.8        & 8.4    & 68.3       &   200.3    &       181.1       \\
            SegNext-T                         & 4.3        & 6.6    & 77.9       &   197.5    &       57.3        \\
            EDAFormer-T                       & 5.0        & 5.6    & 75.9       &   165.0    &       57.8        \\
            CGRSeg-T                          & 9.4        & 4.8    & 78.8       &   135.6    &       359.6       \\
            SegMAN-T                          & 6.4        & 6.0    & 79.0       &   123.8    &       193.0       \\
            FGD2HF-T                          & 5.0        & 5.8    & 79.8       &   96.0     &       188.6       \\
            \hline
            Segformer-B2                      & 27.4       & 60.7   & 78.0       &   59.0     &       596.5       \\
            SegNext-B                         & 27.6       & 34.9   & 82.0       &   101.3    &       189.3       \\
            EDAFormer-B                       & 29.5       & 31.9   & 81.7       &   83.3     &       174.9       \\
            CGRSeg-L                          & 35.7       & 16.5   & 83.4       &   65.6     &       483.4       \\
            SegMAN-S                          & 30.0       & 25.1   & 83.9       &   69.4     &       350.4       \\
            SegAliasing                       & 31.5       & 37.4   & 76.1       &   75.8     &       718.7       \\
            SFFNet                            & 34.2       & 52.3   & 79.9       &   78.7     &       271.5       \\
            FGD2HF-S                          & 23.4       & 19.7   & 83.6       &   71.7     &       299.7       \\
            \hline
            Segformer-B3                      & 47.2       & 77.4   & 80.4       &   46.0     &       672.4       \\
            MaskFormer                        & 42.0       & 55.0   & 81.1       &   133.2    &       863.7       \\
            Mask2Former                       & 47.0       & 74.2   & 82.3       &   48.1     &       845.4       \\
            SegNext-L                         & 48.8       & 70.0   & 82.1       &   59.0     &       281.0       \\
            VWFormer-T                        & 44.8       & 60.0   & 82.9       &   82.7     &       469.9       \\
            SegMAN-B                          & 51.8       & 57.8   & 84.2       &   39.7     &       458.1       \\
            FGD2HF-B                          & 47.2       & 47.2   & 84.0       &   46.4     &       386.8       \\
            \hline
            Segformer-B5                      & 84.6       & 110.0  & 81.0       &   31.9     &       814.4       \\
            VWFormer-S                        & 66.1       & 83.7   & 83.3       &   60.6     &       551.4       \\
            SegMAN-L                          & 92.3       & 97.2   & 84.3       &   27.7     &       587.6       \\
            FsaNet                            & 69.0       & 276.6  & 78.7       &   21.9     &       522.0       \\
            FADC                              & 67.4       & 230.7  & 77.4       &   44.0     &       745.3       \\
            FGD2HF-L                          & 90.8       & 93.5   & 84.7       &   36.3     &       637.4       \\
            \hline
        \end{tabular}
    }
    \label{table5}
\end{table}

\subsection{Analysis of Frequency-guided Dual-domain Fusion}
\begin{figure*}
    \centering
    \includegraphics[width=0.85\textwidth]{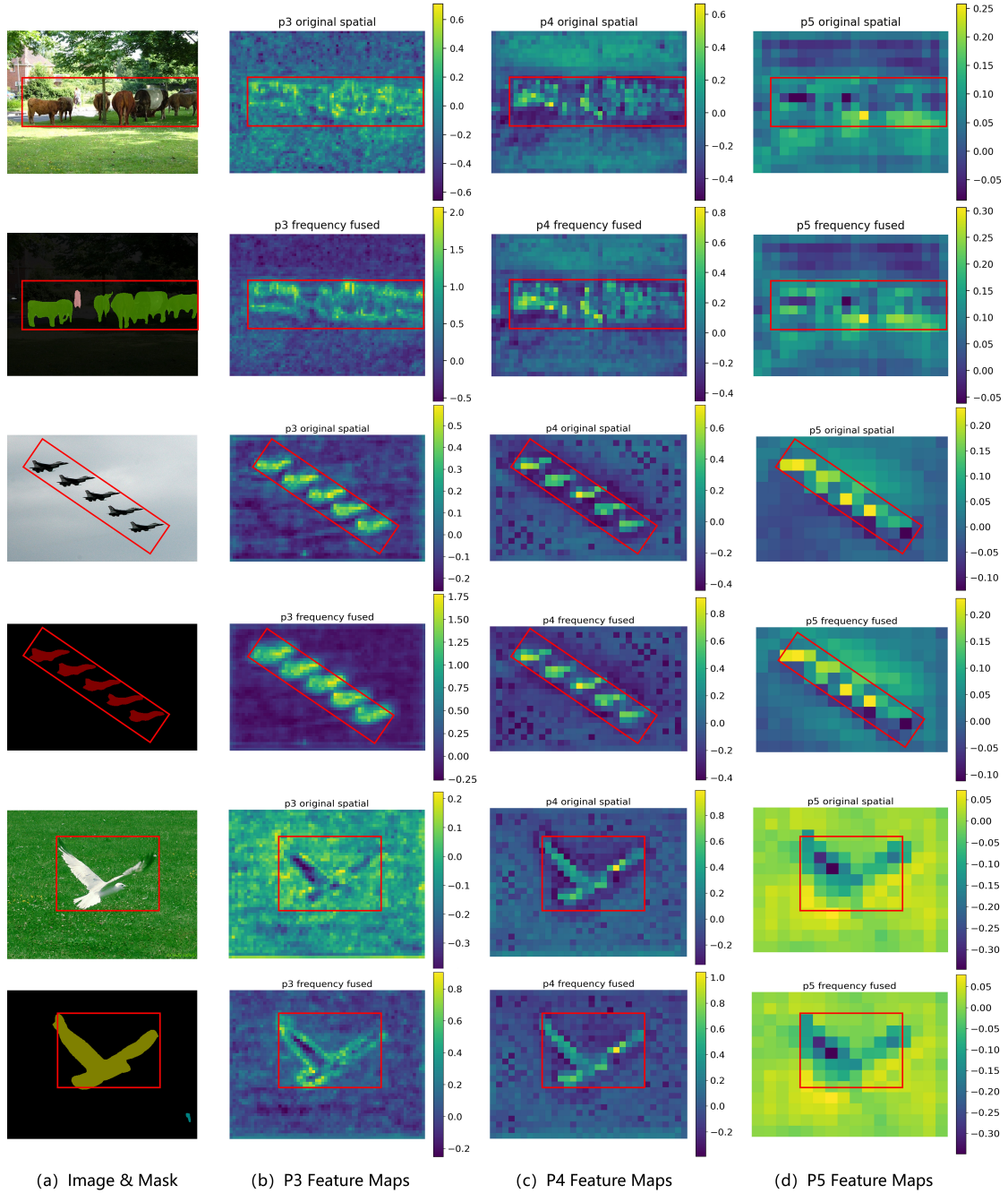}
    \caption{Visualization of feature maps at different stages (P3-P5). (a) Source image and mask; (b) Feature maps of the P3 stage; (c) Feature maps of the P4 stage; (d) Feature maps of the P5 stage. In each group, the first row shows the original feature outputs from the NAHA encoder, and the second row shows the frequency-fusion-enhanced feature outputs from the FGDF module.
}
    \label{fig5}
\end{figure*}

To qualitatively assess the effectiveness of frequency-guided dual-domain fusion in enhancing spatial representations, we visualize the enhanced feature maps at stages P3--P5 and compare them with the corresponding NAHA encoder outputs. As shown in Fig.~\ref{fig5}, the enhanced features exhibit more distinct object details, clearer boundary responses, and greater foreground--background activation contrast. These improvements are particularly evident at P3 and P4, whereas the effect is less pronounced at P5, where foreground and background activations remain similar in the displayed maps. This trend is consistent with the finer spatial detail retained at earlier feature levels. These visualizations provide qualitative support for the ability of frequency guidance to strengthen spatial detail and boundary-related representations, complementing the quantitative segmentation and boundary evaluation results.

We also compare FGDF with a spatial-only hypergraph fusion (SHF) variant. In this variant, spatial features from the preceding stage are downsampled to match the spatial resolution of the current-stage features and used in place of frequency cues to condition node--hyperedge associations. The resulting associations guide the enhancement of current-stage spatial features through hypergraph propagation. As shown in Table~\ref{tablefre}, FGDF achieves an mIoU of 83.6\% and an mBIoU of 77.9\% on SBD, compared with 81.9\% and 73.6\%, respectively, for the spatial-only counterpart. These improvements support the effectiveness of frequency-domain guidance in improving both overall segmentation accuracy and boundary quality.
\begin{table}[t]
    \centering
    \caption{Ablation study of the conditioning source for cross-stage hypergraph fusion on SBD using FGD2HF-S.}
    \label{tablefre}
    \resizebox{0.8\columnwidth}{!}{
        \begin{tabular}{@{}c|ccccc@{}}
            \toprule
            Fusion variant     & mIoU (\%)       & mBIoU (\%)    & mBF1 (\%)     \\
            \midrule
            SHF                & 81.9 (-1.7)     & 73.6 (-4.3)   & 71.9 (-5.1)   \\
            FGDF (ours)        & 83.6            & 77.9          & 77.0          \\
            \bottomrule
        \end{tabular}
    }
\end{table}

\subsection{Ablation Studies on NAHA Encoder}
\label{sec:ablation_naha}
We systematically conduct ablation experiments to verify the effectiveness of NAHA encoder and necessity of its componements. The experiments were conducted on the SBD dataset using the FGD2HF-S variant, and the results are summarized in Table~\ref{table6}.

\textbf{Alternative encoders.} We first design multiple comparative experiments by replacing the NAHA encoder of FGD2HF with other backbones of similar model sizes, including MIT-B2, MSCAN-B, Swin-T, ConvNeXt-T, SegMAN-Encoder-S, and EfficientVMamba-B. These selected backbones cover traditional CNNs, ViTs, and Vim-based architectures, which collectively provide a comprehensive basis to demonstrate the superiority of NAHA. Experimental results demonstrate that our proposed NAHA encoder achieves the best overall performance among all compared backbones, surpassing the second-best competitor (SegMAN-Encoder-S) by 0.4 percentage points, while simultaneously maintaining the lowest parameter count and GFLOPs among all variants. Although its FPS performance is not the highest across all configurations, it still reaches 71.7 Hz, which is sufficient to meet the requirements of most real-world applications.

To further validate the necessity of the architectural design and components choice of NAHA, we also investigate the performance of different variants. It should be noted that, in the NAHA Encoder illustrated in Fig.~\ref{fig3}, each NA Block is simplified to represent the stacking of multiple NA Blocks, which together form one NA layer. This corresponds to the Depths parameter in Table 1.

\textbf{LR-C3AH.} We first examine the impact of the placement of LR-C3AH. Specifically, we compare three configurations: (1) enabling LR-C3AH for global context modeling in every NA Block within each NA layer, (2) disabling LR-C3AH in all NA Blocks, and (3) enabling LR-C3AH only in the last NA Block of each NA layer. Additionally, we also replace LR-C3AH with Linear Attention and 2D Selective Scan (SS2D) to compare different global context modeling approaches. It can be observed that fully enabling LR-C3AH yields the best segmentation performance, with a mean IoU of 83.9\%. However, the test FPS drops considerably, which is mainly attributed to the additional overhead introduced by C3AH's extensive hyperedge construction and hyperedge message passing. In contrast, completely disabling LR-C3AH leads to the worst result (mIoU = 82.2\%). Furthermore, replacing LR-C3AH with SS2D causes a 0.4 percentage-point decrease in mIoU, indicating the effectiveness of our proposed module.

\textbf{HA Block.} To verify the importance of the HA block in Stage 4, we conduct ablation studies by either removing it entirely or replacing it with SS2D or standard attention. The experimental results show that the HA block improves performance by 0.5 and 0.7 percentage points compared with SS2D and standard attention, respectively. Moreover, it achieves a 3.3 percentage point improvement compared with the variant without the HA-based cross-scale global enhancement. These results demonstrate the superiority of HA and highlight the importance of high-order global semantic modeling for accurate semantic segmentation.

\begin{table}[!t]
    \caption{Performance of FGD2HF with Different Encoder Variants on SBD.}
    \centering
    \resizebox{\columnwidth}{!}{
        \begin{tabular}{@{}c|cccc@{}}
            \toprule
            Model Variant                     & Params (M)    & GFLOPs        & mIoU (\%)   & FPS (Hz)            \\
            \midrule
            NAHA                              & 23.4          & 19.7          & 83.6        & 71.7                \\
            \midrule
            \multicolumn{5}{c}{Encoder Replace}                                                                   \\
            \midrule
            MIT-B2                            & 33.9 (+10.5)  & 26.7 (+7.0)   & 80.5 (-3.1) & 78.8 (+7.1)         \\
            MSCAN-B                           & 36.0 (+12.6)  & 33.1 (+13.4)  & 82.7 (-0.9) & 77.1 (+5.4)         \\
            Swin-T                            & 46.4 (+23.0)  & 37.0 (+17.3)  & 81.9 (-1.7) & 80.1 (+8.4)         \\
            ConvNeXt-T                        & 46.7 (+23.3)  & 34.7 (+15.0)  & 81.7 (-1.9) & 87.7 (+16.0)        \\
            SegMAN-Encoder-S                  & 37.3 (+13.9)  & 26.6 (+6.9)   & 83.2 (-0.4) & 56.3 (-15.4)        \\
            EfficientVMamba-B                 & 42.9 (+19.5)  & 26.0 (+6.3)   & 82.9 (-0.7) & 95.3 (+23.6)        \\
            \midrule
            \multicolumn{5}{c}{LR-C3AH Design}                                                                    \\
            \midrule
            All Blocks                        & 27.8 (+4.4)   & 23.2 (+3.5)   & 83.9 (+0.3) & 54.6 (-17.1)        \\
            w/o LR-C3AH                       & 22.5 (-0.9)   & 18.9 (-0.8)   & 82.2 (-1.4) & 82.1 (+10.4)        \\
            Last Block Only (Ours)            & 23.4          & 19.7          & 83.6        & 71.7                \\
            Replace LR-C3AH w SS2D            & 22.6 (-0.8)   & 18.9 (-0.8)   & 83.2 (-0.4) & 72.7 (+1.0)         \\
            Replace LR-C3AH w Linear Attention& 23.2 (-0.2)   & 19.3 (-0.4)   & 82.9 (-0.7) & 73.4 (+1.7)         \\
            \midrule
            \multicolumn{5}{c}{HA Design}                                                                         \\
            \midrule
            Remove HA                         & 22.3 (-1.1)   & 18.6 (-1.1)   & 80.3 (-3.3) & 77.2 (+5.5)         \\
            Replace HA w SS2D                 & 22.6 (-0.8)   & 18.7 (-1.0)   & 83.1 (-0.5) & 74.3 (+2.6)         \\
            Replace HA w Atten                & 23.1 (-0.3)   & 18.8 (-0.9)   & 82.9 (-0.7) & 76.7 (+5.0)         \\
            \bottomrule
        \end{tabular}
    }
    \label{table6}
\end{table}

\subsection{Ablation Studies on WTE and FGDF Modules}
To systematically evaluate the effectiveness of the proposed frequency-guided dual-domain hypergraph fusion and enhancement method, we conduct ablation studies on the WTE module and FGDF module. Specifically, the following configurations are compared: (1) removing the them and directly using the spatial features extracted by the NAHA encoder; (2) suppressing the high-frequency information and performing semantic segmentation using only the low-frequency fused and enhanced features; (3) Using the low- and high-frequency features obtained by wavelet decomposition without frequency enhancement; (4) Enhancing both low- and high-frequency features without using low-frequency information to modulate high-frequency responses; and (5) replacing the proposed frequency-guided hypergraph fusion mechanism with alternative spatial-frequency fusion strategies, including Fast Fourier Convolution (FFC) \cite{chi2020fast}, Spatial-Frequency Cross-Attention (SFCA) \cite{gu2023adafuse}, and Gated Spatial-Frequency Fusion (GSFF) \cite{canitez2026frequency}. The ablation experiments are conducted using the FGD2HF-S model on ADE20K, and the results are reported in Table~\ref{table7}.

As shown in Table~\ref{table7}, removing both WTE and FGDF decreases mIoU and mBIoU by 3.8 and 4.4 percentage points, respectively, indicating the contribution of the two modules to overall segmentation and boundary quality. Suppressing high-frequency information reduces mIoU and mBIoU by 0.9 and 2.1 percentage points, respectively, while removing frequency enhancement leads to decreases of 1.2 and 1.7 percentage points. Furthermore, removing low-frequency guidance from high-frequency enhancement decreases mIoU by 0.6 percentage points and mBIoU by 1.4 percentage points, supporting the benefit of low-frequency guidance for boundary preservation. Compared with FFC, SFCA, and GSFF, FGDF achieves higher mIoU and mBIoU with fewer parameters and GFLOPs under the evaluated configurations.

\begin{figure}
    \centering
    \includegraphics[width=0.9\columnwidth]{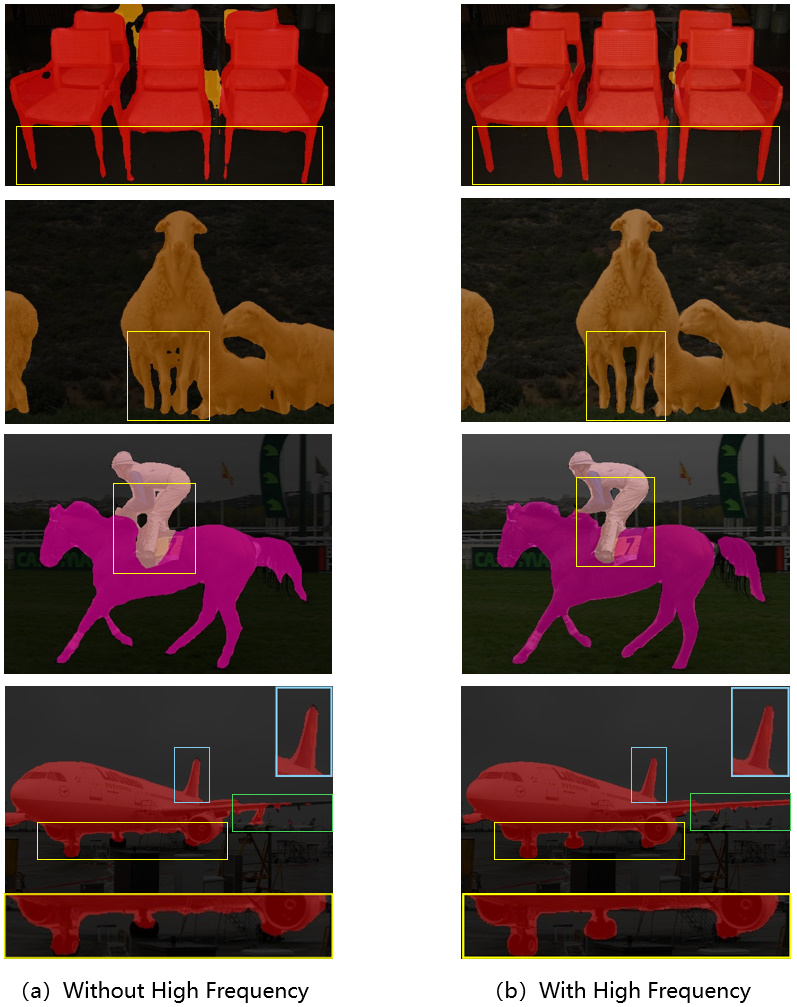}
    \caption{Segmentation results with and without high-frequency components. (a) Without high-frequency components; (b) With high-frequency components. Zoom in for best view.}
    \label{fig6}
\end{figure}

Additionally, Fig.~\ref{fig6} presents a qualitative comparison of segmentation results with and without high-frequency information. The examples show clearer object boundaries when high-frequency information is incorporated, complementing the quantitative boundary evaluation.

\begin{table}[!t]
    \caption{Ablation Results the WTE and FGDF modules on the ADE20K Dataset.}
    \centering
    \resizebox{0.9\columnwidth}{!}{
        \begin{tabular}{@{}c|cccc@{}}
            \toprule
            Model Variant                                   & Params (M)       & GFLOPs       & mIoU (\%)      & mBIoU (\%)   \\
            \midrule
            FGD2HF                                          & 23.4             & 19.7         & 48.1           & 38.6         \\
            FGD2HF w/o WTE\&FGDF                            & 18.1 (-5.3)      & 16.5 (-3.2)  & 44.3 (-3.8)    & 34.2 (-4.4)  \\
            \midrule
            \multicolumn{4}{c}{Information Suppress}                                                                          \\
            \midrule
            Suppress high-fre info                          & 23.4             & 19.7         & 47.2 (-0.9)    & 36.5 (-2.1)  \\
            WTE w/o enhancement                             & 22.1  (-1.3)     & 18.9 (-0.8)  & 46.9 (-1.2)    & 36.9 (-1.7)  \\
            WTE w/o low-fre modulation                      & 23.2  (-0.2)     & 19.6 (-0.1)  & 47.5 (-0.6)    & 37.2 (-1.4)  \\
            \midrule
            \multicolumn{4}{c}{Fusion Mechanism Replace}                                                                      \\
            \midrule
            Replace FGDF w FFC  \cite{chi2020fast}          & 25.6  (+2.2)     & 22.1  (+2.4) & 46.8 (-1.3)    &  35.8 (-2.8) \\
            Replace FGDF w SFCA \cite{gu2023adafuse}        & 24.8  (+1.4)     & 23.7  (+4.0) & 47.0 (-1.1)    &  36.8 (-1.8) \\
            Replace FGDF w GSFF \cite{canitez2026frequency} & 26.1  (+2.7)     & 24.3  (+4.6) & 47.4 (-0.7)    &  37.0 (-1.6) \\
            \bottomrule
        \end{tabular}
    }
    \label{table7}
\end{table}

\subsection{Ablation Studies on FGD2HF Decoder}
We evaluate the overall effectiveness of the proposed decoder on the SBD dataset. For a fair comparison, all model variants are built upon the FGD2HF-S encoder, and the number of channels in different decoders is adjusted to ensure comparable parameters across different decoder designs. The experimental results are presented in Table~\ref{table8}. Our hyper decoder achieves the highest segmentation accuracy while maintaining the lowest computational cost, at just $19.7$ GFLOPs. More notably, when integrated with our FGD2HF-S encoder, the SegFormer, LightHam, and EDAFormer decoders achieve mIoU improvements of $4.4$, $0.6$, and $1.4$ percentage points, respectively, over their original architectures, namely SegFormer, SegNeXt, and EDAFormer. These consistent improvements firmly validate the effectiveness of our proposed NAHA and frequency-guided dual-domain hypergraph fusion modules for semantic segmentation tasks.
\begin{table}[!t]
    \caption{Performance of FGD2HF with Different Decoders on SBD.}
    \centering
    \resizebox{\columnwidth}{!}{
        \begin{tabular}{@{}c|cccc@{}}
            \hline
            Model Variant                             & Params (M)       & GFLOPs            & mIoU (\%)        & FPS (Hz)         \\
            \hline
            FCN Decoder  \cite{long2015fully}         & 24.6 (+1.2)      & 26.9 (+7.2)       & 81.7 (-1.9)      & 76.4 (+4.7)      \\
            SegFormer Decoder \cite{xie2021segformer} & 23.1 (-0.3)      & 23.5 (+3.8)       & 82.4 (-1.2)      & 74.1 (+2.4)      \\
            LightHam Decoder  \cite{guo2022segnext}   & 23.0 (-0.4)      & 20.1 (+0.4)       & 82.6 (-1.0)      & 69.2 (-2.5)      \\
            EDAFormer Decoder \cite{yu2024embedding}  & 26.3 (+2.9)      & 20.7 (+1.0)       & 83.1 (-0.5)      & 70.7 (-1.0)      \\
            SegMAN Decoder  \cite{fu2025segman}       & 28.0 (+4.6)      & 22.4 (+2.7)       & 83.3 (-0.3)      & 71.5 (-0.2)      \\
            Hyper Decoder (Ours)                      & 23.4             & 19.7              & 83.6             & 71.7             \\
            \hline
        \end{tabular}
    }
    \label{table8}
\end{table}

\section{Conclusion and Future Work}
This article presents FGD2HF, a novel dual-domain fusion network that enhances semantic segmentation by leveraging low- and high-frequency components to fuse and enhance spatial-domain features via frequency-guided hypergraph attention. Our approach includes a NAHA encoder that extracts multi-level features using neighborhood and hypergraph attention, boosting the feature extraction capability of FGD2HF. Through the integration of the WTE module and FGDF module, FGD2HF effectively captures both global semantics and local texture details by fusing low- and high-frequency information. Wavelet transform is employed to extract frequency information from the preceding stage, which is then used to enhance spatial information fusion in the subsequent stage. This mechanism effectively alleviates the over-smoothing phenomenon commonly caused by the loss of high-frequency information in deeper network layers, while maintaining low computational complexity. 

Extensive experiments on multiple public datasets demonstrate that FGD2HF achieves highly competitive or superior performance across multiple benchmark datasets. The frequency-guided dual-domain hypergraph fusion strategy effectively improves semantic segmentation accuracy, especially in complex scenes. Furthermore, FGD2HF achieves superior accuracy with lower computational resource requirements, demonstrating an excellent trade-off between segmentation accuracy and computational efficiency.

Future work will focus on further optimizing FGD2HF, including reducing network parameters and model size without sacrificing performance, as well as improving its inference speed. We will also explore the deep integration of other advanced model architectures with FGD2HF to further enhance its segmentation capabilities.

\appendix
\section{Segmentation Examples}
\begin{figure*}
    \centering
    \includegraphics[width=\textwidth]{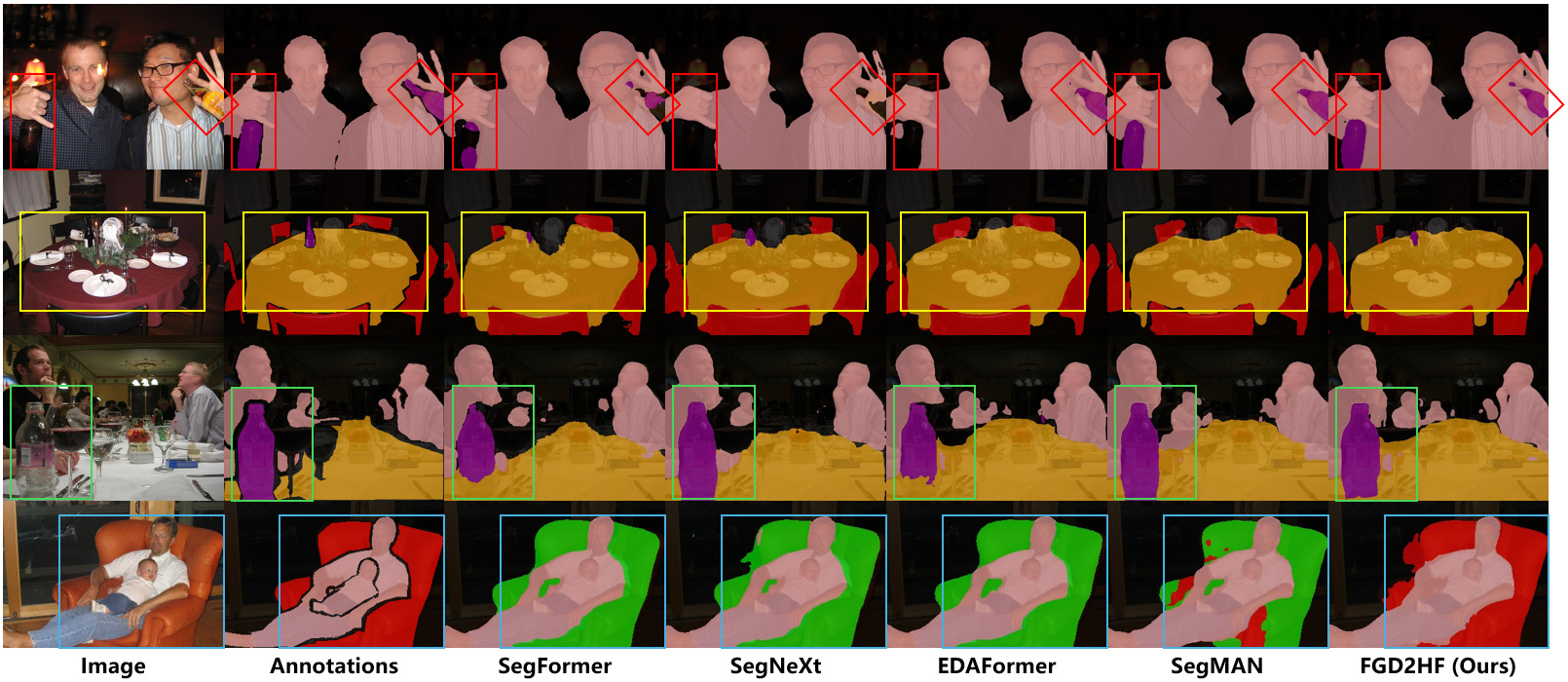}
    \caption{Segmentation examples on SBD. Zoom in for best view.}
    \label{fig7}
\end{figure*}

\begin{figure*}
    \centering
    \includegraphics[width=\textwidth]{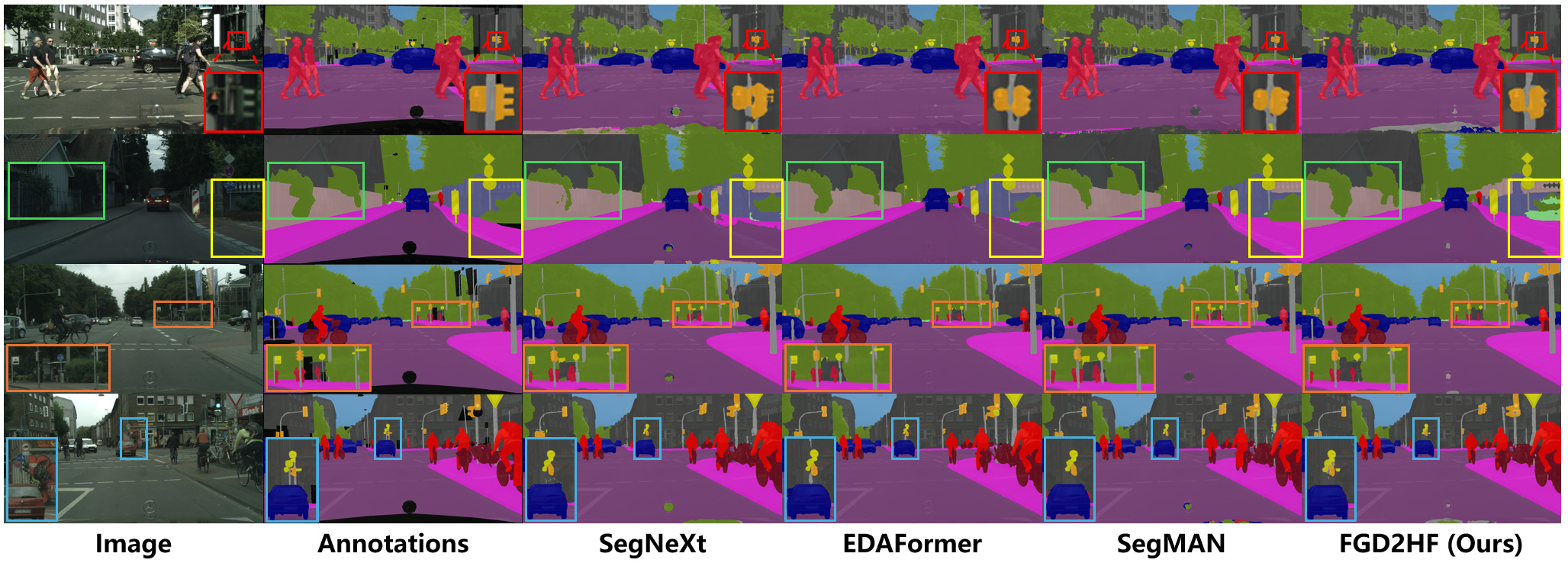}
    \caption{Segmentation examples on Cityscapes. Zoom in for best view.}
    \label{fig8}
\end{figure*}

\begin{figure*}
    \centering
    \includegraphics[width=\textwidth]{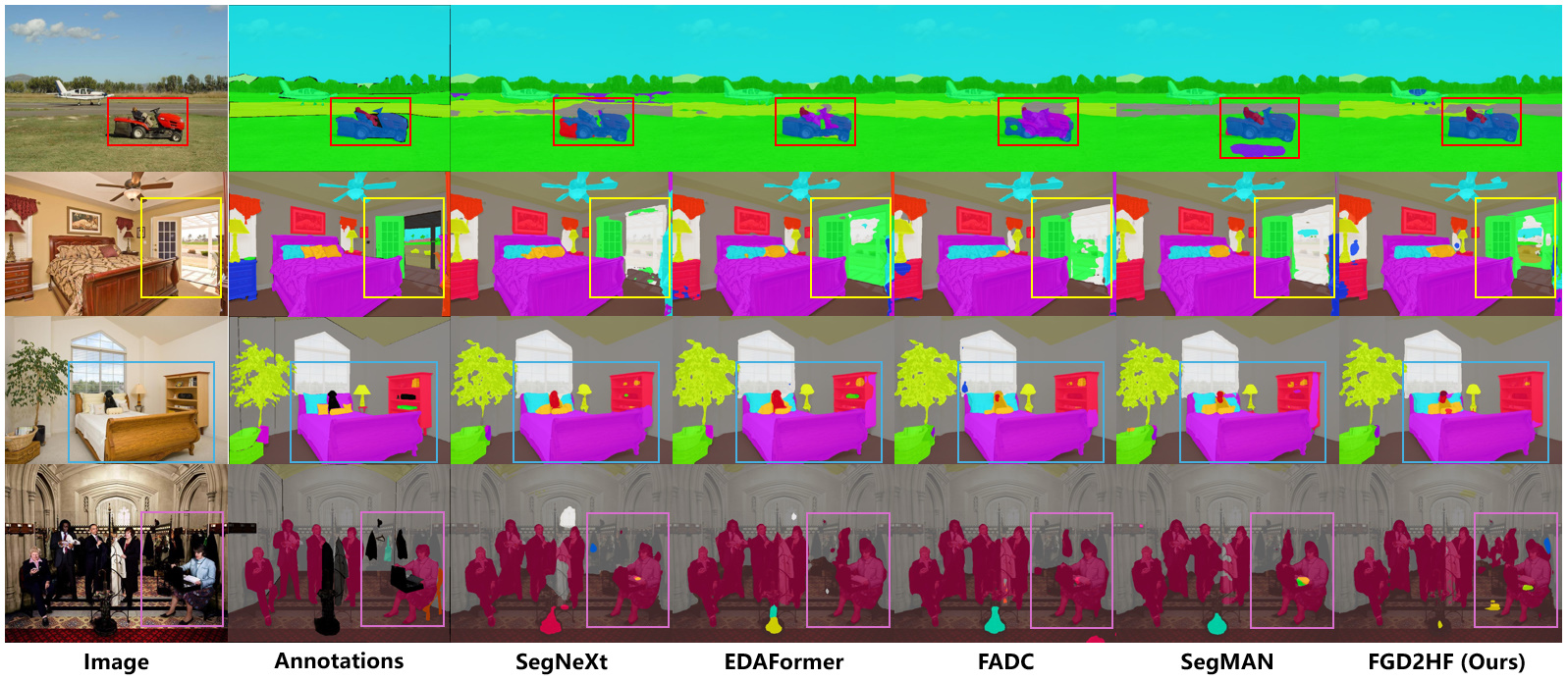}
    \caption{Segmentation examples on ADE20K. Zoom in for best view.}
    \label{fig9}
\end{figure*}

Fig.~\ref{fig7}, \ref{fig8}, and \ref{fig9} illustrate the segmentation examples of different models on the SBD, Cityscapes, and ADE20K datasets, respectively. As can be observed from these qualitative examples, our proposed FGD2HF method, which leverages frequency-based hypergraph fusion, is more effective at capturing both fine-grained local dependencies and long-range contextual information. Consequently, it exhibits cleaner boundaries and improved semantic coherence in the selected examples.

\printcredits

\bibliographystyle{cas-model2-names}
\bibliography{reference}

@article{li2026dyfuse,
  title={Dyfuse: Dynamic fusion for weakly-supervised semantic segmentation in autonomous driving scenes},
  author={Li, Yongqiang and Hu, Chuanping and Ren, Kai and Xi, Hao and Fan, Jinhao},
  journal={Information Fusion},
  volume={125},
  pages={103471},
  year={2026},
  publisher={Elsevier}
}

@article{wang2025reliable,
  title={Reliable autonomous driving environment perception: uncertainty quantification of semantic segmentation},
  author={Wang, Rui and Yang, Tengkun and Liang, Ci and Wang, Mengying and Ci, Yusheng},
  journal={Journal of Transportation Engineering, Part A: Systems},
  volume={151},
  number={3},
  pages={04024117},
  year={2025},
  publisher={American Society of Civil Engineers}
}

@article{silva2025exploring,
  title={Exploring the effects of synthetic data generation: a case study on autonomous driving for semantic segmentation: M. Silva et al.},
  author={Silva, Manuel and Seoane, Antonio and Mures, Omar A and Lopez, Antonio M and Iglesias-Guitian, Jose A},
  journal={The visual computer},
  volume={41},
  number={10},
  pages={7379--7397},
  year={2025},
  publisher={Springer}
}

@inproceedings{yang2025hsrdiff,
  title={Hsrdiff: A hierarchical self-regulation diffusion model for stochastic semantic segmentation},
  author={Yang, Han and Yang, Chuanguang and An, Zhulin and Huang, Libo and Xu, Yongjun},
  booktitle={Proceedings of the AAAI Conference on Artificial Intelligence},
  volume={39},
  number={2},
  pages={1701--1709},
  year={2025}
}

@article{krithika2022review,
  title={Review of semantic segmentation of medical images using modified architectures of UNET},
  author={Krithika Alias AnbuDevi, M and Suganthi, K},
  journal={Diagnostics},
  volume={12},
  number={12},
  pages={3064},
  year={2022},
  publisher={MDPI}
}

@article{singh2023semantic,
  title={Semantic segmentation of satellite images using deep-unet},
  author={Singh, Ningthoujam Johny and Nongmeikapam, Kishorjit},
  journal={Arabian Journal for Science and Engineering},
  volume={48},
  number={2},
  pages={1193--1205},
  year={2023},
  publisher={Springer}
}

@article{amirgan2024semantic,
  title={Semantic segmentation of satellite images with different building types using deep learning methods},
  author={Amirgan, Burcu and Erener, Arzu},
  journal={Remote Sensing Applications: Society and Environment},
  volume={34},
  pages={101176},
  year={2024},
  publisher={Elsevier}
}

@article{zhong2024novel,
  title={A novel grasp detection algorithm with multi-target semantic segmentation for a robot to manipulate cluttered objects},
  author={Zhong, Xungao and Chen, Yijun and Luo, Jiaguo and Shi, Chaoquan and Hu, Huosheng},
  journal={Machines},
  volume={12},
  number={8},
  pages={506},
  year={2024},
  publisher={MDPI}
}

@article{brunke2025semantically,
  title={Semantically safe robot manipulation: From semantic scene understanding to motion safeguards},
  author={Brunke, Lukas and Zhang, Yanni and R{\"o}mer, Ralf and Naimer, Jack and Staykov, Nikola and Zhou, Siqi and Schoellig, Angela P},
  journal={IEEE Robotics and Automation Letters},
  year={2025},
  publisher={IEEE}
}

@article{jiang2024geometry,
  title={Geometry perception and motion planning in robotic assembly based on semantic segmentation and point clouds reconstruction},
  author={Jiang, Yuze and Liu, Guanghui and Huang, Zhouzhou and Yang, Bin and Yang, Wenyu},
  journal={Engineering Applications of Artificial Intelligence},
  volume={130},
  pages={107678},
  year={2024},
  publisher={Elsevier}
}

@inproceedings{long2015fully,
  title={Fully convolutional networks for semantic segmentation},
  author={Long, Jonathan and Shelhamer, Evan and Darrell, Trevor},
  booktitle={Proceedings of the IEEE conference on computer vision and pattern recognition},
  pages={3431--3440},
  year={2015}
}

@inproceedings{he2017mask,
  title={Mask r-cnn},
  author={He, Kaiming and Gkioxari, Georgia and Doll{\'a}r, Piotr and Girshick, Ross},
  booktitle={Proceedings of the IEEE international conference on computer vision},
  pages={2961--2969},
  year={2017}
}

@inproceedings{ronneberger2015u,
  title={U-net: Convolutional networks for biomedical image segmentation},
  author={Ronneberger, Olaf and Fischer, Philipp and Brox, Thomas},
  booktitle={International Conference on Medical image computing and computer-assisted intervention},
  pages={234--241},
  year={2015},
  organization={Springer}
}

@article{chen2017deeplab,
  title={Deeplab: Semantic image segmentation with deep convolutional nets, atrous convolution, and fully connected crfs},
  author={Chen, Liang-Chieh and Papandreou, George and Kokkinos, Iasonas and Murphy, Kevin and Yuille, Alan L},
  journal={IEEE transactions on pattern analysis and machine intelligence},
  volume={40},
  number={4},
  pages={834--848},
  year={2017},
  publisher={IEEE}
}

@article{dosovitskiy2020image,
  title={An image is worth 16x16 words: Transformers for image recognition at scale},
  author={Dosovitskiy, Alexey and Beyer, Lucas and Kolesnikov, Alexander and Weissenborn, Dirk and Zhai, Xiaohua and Unterthiner, Thomas and Dehghani, Mostafa and Minderer, Matthias and Heigold, Georg and Gelly, Sylvain and others},
  journal={arXiv preprint arXiv:2010.11929},
  year={2020}
}

@inproceedings{strudel2021segmenter,
  title={Segmenter: Transformer for semantic segmentation},
  author={Strudel, Robin and Garcia, Ricardo and Laptev, Ivan and Schmid, Cordelia},
  booktitle={Proceedings of the IEEE/CVF international conference on computer vision},
  pages={7262--7272},
  year={2021}
}

@inproceedings{zheng2021rethinking,
  title={Rethinking semantic segmentation from a sequence-to-sequence perspective with transformers},
  author={Zheng, Sixiao and Lu, Jiachen and Zhao, Hengshuang and Zhu, Xiatian and Luo, Zekun and Wang, Yabiao and Fu, Yanwei and Feng, Jianfeng and Xiang, Tao and Torr, Philip HS and others},
  booktitle={Proceedings of the IEEE/CVF conference on computer vision and pattern recognition},
  pages={6881--6890},
  year={2021}
}

@article{xie2021segformer,
  title={SegFormer: Simple and efficient design for semantic segmentation with transformers},
  author={Xie, Enze and Wang, Wenhai and Yu, Zhiding and Anandkumar, Anima and Alvarez, Jose M and Luo, Ping},
  journal={Advances in neural information processing systems},
  volume={34},
  pages={12077--12090},
  year={2021}
}

@inproceedings{chen2021dpt,
  title={Dpt: Deformable patch-based transformer for visual recognition},
  author={Chen, Zhiyang and Zhu, Yousong and Zhao, Chaoyang and Hu, Guosheng and Zeng, Wei and Wang, Jinqiao and Tang, Ming},
  booktitle={Proceedings of the 29th ACM international conference on multimedia},
  pages={2899--2907},
  year={2021}
}

@inproceedings{wang2021pyramid,
  title={Pyramid vision transformer: A versatile backbone for dense prediction without convolutions},
  author={Wang, Wenhai and Xie, Enze and Li, Xiang and Fan, Deng-Ping and Song, Kaitao and Liang, Ding and Lu, Tong and Luo, Ping and Shao, Ling},
  booktitle={Proceedings of the IEEE/CVF international conference on computer vision},
  pages={568--578},
  year={2021}
}

@article{cheng2021per,
  title={Per-pixel classification is not all you need for semantic segmentation},
  author={Cheng, Bowen and Schwing, Alex and Kirillov, Alexander},
  journal={Advances in neural information processing systems},
  volume={34},
  pages={17864--17875},
  year={2021}
}

@inproceedings{cheng2022masked,
  title={Masked-attention mask transformer for universal image segmentation},
  author={Cheng, Bowen and Misra, Ishan and Schwing, Alexander G and Kirillov, Alexander and Girdhar, Rohit},
  booktitle={Proceedings of the IEEE/CVF conference on computer vision and pattern recognition},
  pages={1290--1299},
  year={2022}
}

@article{chen2023spatialfrequency,
  title={Spatial Frequency Modulation for Semantic Segmentation},
  author={Chen, Linwei and Fu, Ying and Gu, Lin and Zheng, Dezhi and Dai, Jifeng},
  journal={IEEE Transactions on Pattern Analysis and Machine Intelligence},
  year={2025},
  volume={47},
  number={11},
  pages={9767-9784},
  doi={10.1109/TPAMI.2025.3592621}
}

@article{bi2024learning,
  title={Learning frequency-adapted vision foundation model for domain generalized semantic segmentation},
  author={Bi, Qi and Yi, Jingjun and Zheng, Hao and Zhan, Haolan and Huang, Yawen and Ji, Wei and Li, Yuexiang and Zheng, Yefeng},
  journal={Advances in Neural Information Processing Systems},
  volume={37},
  pages={94047--94072},
  year={2024}
}

@inproceedings{chen2024semantic,
  title={When Semantic Segmentation Meets Frequency Aliasing},
  author={Chen, Linwei and Gu, Lin and Fu, Ying},
  booktitle={International Conference on Learning Representations},
  year={2024}
}

@article{chen2017rethinking,
  title={Rethinking atrous convolution for semantic image segmentation},
  author={Chen, Liang-Chieh and Papandreou, George and Schroff, Florian and Adam, Hartwig},
  journal={arXiv preprint arXiv:1706.05587},
  year={2017}
}

@inproceedings{chen2018encoder,
  title={Encoder-decoder with atrous separable convolution for semantic image segmentation},
  author={Chen, Liang-Chieh and Zhu, Yukun and Papandreou, George and Schroff, Florian and Adam, Hartwig},
  booktitle={Proceedings of the European conference on computer vision (ECCV)},
  pages={801--818},
  year={2018}
}

@article{thisanke2023semantic,
  title={Semantic segmentation using vision transformers: A survey},
  author={Thisanke, Hans and Deshan, Chamli and Chamith, Kavindu and Seneviratne, Sachith and Vidanaarachchi, Rajith and Herath, Damayanthi},
  journal={Engineering Applications of Artificial Intelligence},
  volume={126},
  pages={106669},
  year={2023},
  publisher={Elsevier}
}

@article{aslan2026frequency,
  title={Frequency-Domain Vision Transformers: Architectures, Applications, and Open Challenges},
  author={Aslan, Muhammet Fatih and Aslan, Busra and Sabanci, Kadir},
  journal={Applied Sciences},
  volume={16},
  number={4},
  pages={2024},
  year={2026},
  publisher={MDPI}
}

@inproceedings{ni2024context,
  title={Context-guided spatial feature reconstruction for efficient semantic segmentation},
  author={Ni, Zhenliang and Chen, Xinghao and Zhai, Yingjie and Tang, Yehui and Wang, Yunhe},
  booktitle={European conference on computer vision},
  pages={239--255},
  year={2024},
  organization={Springer}
}

@article{li2025frequency,
  title={A frequency decoupling network for semantic segmentation of remote sensing images},
  author={Li, Xin and Xu, Feng and Yu, Anzhu and Lyu, Xin and Gao, Hongmin and Zhou, Jun},
  journal={IEEE Transactions on Geoscience and Remote Sensing},
  volume={63},
  pages={1--21},
  year={2025},
  publisher={IEEE}
}

@inproceedings{lo2019exploring,
  title={Exploring semantic segmentation on the dct representation},
  author={Lo, Shao-Yuan and Hang, Hsueh-Ming},
  booktitle={Proceedings of the 1st ACM International Conference on Multimedia in Asia},
  pages={1--6},
  year={2019}
}

@inproceedings{yang2020fda,
  title={Fda: Fourier domain adaptation for semantic segmentation},
  author={Yang, Yanchao and Soatto, Stefano},
  booktitle={Proceedings of the IEEE/CVF conference on computer vision and pattern recognition},
  pages={4085--4095},
  year={2020}
}

@inproceedings{chen2025frequency,
  title={Frequency dynamic convolution for dense image prediction},
  author={Chen, Linwei and Gu, Lin and Li, Liang and Yan, Chenggang and Fu, Ying},
  booktitle={Proceedings of the Computer Vision and Pattern Recognition Conference},
  pages={30178--30188},
  year={2025}
}

@article{zhang2023fsanet,
  title={FsaNet: Frequency self-attention for semantic segmentation},
  author={Zhang, Fengyu and Panahi, Ashkan and Gao, Guangjun},
  journal={IEEE Transactions on Image Processing},
  volume={32},
  pages={4757--4772},
  year={2023},
  publisher={IEEE}
}

@inproceedings{chen2024frequency,
  title={Frequency-adaptive dilated convolution for semantic segmentation},
  author={Chen, Linwei and Gu, Lin and Zheng, Dezhi and Fu, Ying},
  booktitle={Proceedings of the IEEE/CVF Conference on Computer Vision and Pattern Recognition},
  pages={3414--3425},
  year={2024}
}

@inproceedings{yang2025ffr,
  title={Ffr: Frequency feature rectification for weakly supervised semantic segmentation},
  author={Yang, Ziqian and Zhao, Xinqiao and Wang, Xiaolei and Zhang, Quan and Xiao, Jimin},
  booktitle={Proceedings of the Computer Vision and Pattern Recognition Conference},
  pages={30261--30270},
  year={2025}
}

@article{li2025dual,
  title={Dual-domain decoupled fusion network for semantic segmentation of remote sensing images},
  author={Li, Xin and Xu, Feng and Zhang, Jue and Yu, Anzhu and Lyu, Xin and Gao, Hongmin and Zhou, Jun},
  journal={Information Fusion},
  volume={124},
  pages={103359},
  year={2025},
  publisher={Elsevier}
}

@article{yang2024sffnet,
  title={SFFNet: A wavelet-based spatial and frequency domain fusion network for remote sensing segmentation},
  author={Yang, Yunsong and Yuan, Genji and Li, Jinjiang},
  journal={IEEE Transactions on Geoscience and Remote Sensing},
  volume={62},
  pages={1--17},
  year={2024},
  publisher={IEEE}
}

@article{badrinarayanan2017segnet,
  title={Segnet: A deep convolutional encoder-decoder architecture for image segmentation},
  author={Badrinarayanan, Vijay and Kendall, Alex and Cipolla, Roberto},
  journal={IEEE transactions on pattern analysis and machine intelligence},
  volume={39},
  number={12},
  pages={2481--2495},
  year={2017},
  publisher={IEEE}
}

@article{younesi2024comprehensive,
  title={A comprehensive survey of convolutions in deep learning: Applications, challenges, and future trends},
  author={Younesi, Abolfazl and Ansari, Mohsen and Fazli, Mohammadamin and Ejlali, Alireza and Shafique, Muhammad and Henkel, J{\"o}rg},
  journal={IEEE Access},
  volume={12},
  pages={41180--41218},
  year={2024},
  publisher={IEEE}
}

@article{csurka2022semantic,
  title={Semantic image segmentation: Two decades of research},
  author={Csurka, Gabriela and Volpi, Riccardo and Chidlovskii, Boris},
  journal={Foundations and Trends in Computer Graphics and Vision},
  volume={14},
  number={1-2},
  pages={1--162},
  year={2022},
  publisher={Emerald Publishing Limited}
}

@article{arbelaez2010contour,
  title={Contour detection and hierarchical image segmentation},
  author={Arbelaez, Pablo and Maire, Michael and Fowlkes, Charless and Malik, Jitendra},
  journal={IEEE transactions on pattern analysis and machine intelligence},
  volume={33},
  number={5},
  pages={898--916},
  year={2010},
  publisher={IEEE}
}

@article{wang2010generalizing,
  title={Generalizing edge detection to contour detection for image segmentation},
  author={Wang, Hongzhi and Oliensis, John},
  journal={Computer Vision and Image Understanding},
  volume={114},
  number={7},
  pages={731--744},
  year={2010},
  publisher={Elsevier}
}

@article{comaniciu2002mean,
  title={Mean shift: A robust approach toward feature space analysis},
  author={Comaniciu, Dorin and Meer, Peter},
  journal={IEEE Transactions on pattern analysis and machine intelligence},
  volume={24},
  number={5},
  pages={603--619},
  year={2002},
  publisher={IEEE}
}

@article{achanta2012slic,
  title={SLIC superpixels compared to state-of-the-art superpixel methods},
  author={Achanta, Radhakrishna and Shaji, Appu and Smith, Kevin and Lucchi, Aurelien and Fua, Pascal and S{\"u}sstrunk, Sabine},
  journal={IEEE transactions on pattern analysis and machine intelligence},
  volume={34},
  number={11},
  pages={2274--2282},
  year={2012},
  publisher={IEEE}
}

@article{wang2011color,
  title={Color image segmentation using pixel wise support vector machine classification},
  author={Wang, Xiang-Yang and Wang, Ting and Bu, Juan},
  journal={Pattern Recognition},
  volume={44},
  number={4},
  pages={777--787},
  year={2011},
  publisher={Elsevier}
}

@article{pal2005support,
  title={Support vector machines for classification in remote sensing},
  author={Pal, Mahesh and Mather, Paul M},
  journal={International journal of remote sensing},
  volume={26},
  number={5},
  pages={1007--1011},
  year={2005},
  publisher={Taylor \& Francis}
}

@article{geman1984stochastic,
  title={Stochastic relaxation, Gibbs distributions, and the Bayesian restoration of images},
  author={Geman, Stuart and Geman, Donald},
  journal={IEEE Transactions on pattern analysis and machine intelligence},
  number={6},
  pages={721--741},
  year={1984},
  publisher={IEEE}
}

@article{shotton2009textonboost,
  title={Textonboost for image understanding: Multi-class object recognition and segmentation by jointly modeling texture, layout, and context},
  author={Shotton, Jamie and Winn, John and Rother, Carsten and Criminisi, Antonio},
  journal={International journal of computer vision},
  volume={81},
  number={1},
  pages={2--23},
  year={2009},
  publisher={Springer}
}

@article{stuchi2024frequency,
  title={A frequency-domain approach with learnable filters for image classification},
  author={Stuchi, Jos{\'e} Augusto and Canto, Natalia Gil and de Faissol Attux, Romis Ribeiro and Boccato, Levy},
  journal={Applied Soft Computing},
  volume={155},
  pages={111443},
  year={2024},
  publisher={Elsevier}
}

@article{hou2022multisignal,
  title={Multisignal modulation classification using sliding window detection and complex convolutional network in frequency domain},
  author={Hou, Changbo and Liu, Guowei and Tian, Qiao and Zhou, Zhichao and Hua, Lijie and Lin, Yun},
  journal={IEEE Internet of Things Journal},
  volume={9},
  number={19},
  pages={19438--19449},
  year={2022},
  publisher={IEEE}
}

@article{ling2025fdl,
  title={FDL-YOLO: efficient real-time traffic detection network based on frequency domain learning},
  author={Ling, Hanhuang and Li, Ling and Pan, Duotao and Zhang, Yuxiang},
  journal={The Journal of Supercomputing},
  volume={81},
  number={8},
  pages={966},
  year={2025},
  publisher={Springer}
}

@inproceedings{xu2020learning,
  title={Learning in the frequency domain},
  author={Xu, Kai and Qin, Minghai and Sun, Fei and Wang, Yuhao and Chen, Yen-Kuang and Ren, Fengbo},
  booktitle={Proceedings of the IEEE/CVF conference on computer vision and pattern recognition},
  pages={1740--1749},
  year={2020}
}

@inproceedings{zhong2022detecting,
  title={Detecting camouflaged object in frequency domain},
  author={Zhong, Yijie and Li, Bo and Tang, Lv and Kuang, Senyun and Wu, Shuang and Ding, Shouhong},
  booktitle={Proceedings of the IEEE/CVF conference on computer vision and pattern recognition},
  pages={4504--4513},
  year={2022}
}

@inproceedings{tan2024frequency,
  title={Frequency-aware deepfake detection: Improving generalizability through frequency space domain learning},
  author={Tan, Chuangchuang and Zhao, Yao and Wei, Shikui and Gu, Guanghua and Liu, Ping and Wei, Yunchao},
  booktitle={Proceedings of the AAAI Conference on Artificial Intelligence},
  volume={38},
  number={5},
  pages={5052--5060},
  year={2024}
}

@article{bretto2013hypergraph,
  title={Hypergraph theory},
  author={Bretto, Alain},
  journal={An introduction. Mathematical Engineering. Cham: Springer},
  volume={1},
  pages={209--216},
  year={2013},
  publisher={Springer}
}

@article{li2023efficient,
  title={Efficient and effective attributed hypergraph clustering via k-nearest neighbor augmentation},
  author={Li, Yiran and Yang, Renchi and Shi, Jieming},
  journal={Proceedings of the ACM on Management of Data},
  volume={1},
  number={2},
  pages={1--23},
  year={2023},
  publisher={ACM New York, NY, USA}
}

@article{feng2023modularity,
  title={Modularity-based hypergraph clustering: random hypergraph model, hyperedge-cluster relation, and computation},
  author={Feng, Zijin and Qiao, Miao and Cheng, Hong},
  journal={Proceedings of the ACM on Management of Data},
  volume={1},
  number={3},
  pages={1--25},
  year={2023},
  publisher={ACM New York, NY, USA}
}

@article{guo2025mmhcl,
  title={Mmhcl: Multi-modal hypergraph contrastive learning for recommendation},
  author={Guo, Xu and Zhang, Tong and Wang, Fuyun and Wang, Xudong and Zhang, Xiaoya and Liu, Xin and Cui, Zhen},
  journal={ACM Transactions on Multimedia Computing, Communications and Applications},
  volume={21},
  number={10},
  pages={1--23},
  year={2025},
  publisher={ACM New York, NY}
}

@article{khan2025heterogeneous,
  title={Heterogeneous hypergraph neural network for social recommendation using attention network},
  author={Khan, Bilal and Wu, Jia and Yang, Jian and Ma, Xiaoxiao},
  journal={ACM Transactions on Recommender Systems},
  volume={3},
  number={3},
  pages={1--22},
  year={2025},
  publisher={ACM New York, NY}
}

@inproceedings{kim2020hypergraph,
  title={Hypergraph attention networks for multimodal learning},
  author={Kim, Eun-Sol and Kang, Woo Young and On, Kyoung-Woon and Heo, Yu-Jung and Zhang, Byoung-Tak},
  booktitle={Proceedings of the IEEE/CVF conference on computer vision and pattern recognition},
  pages={14581--14590},
  year={2020}
}

@article{lei2025yolov13,
  title={Yolov13: Real-time object detection with hypergraph-enhanced adaptive visual perception},
  author={Lei, Mengqi and Li, Siqi and Wu, Yihong and Hu, Han and Zhou, You and Zheng, Xinhu and Ding, Guiguang and Du, Shaoyi and Wu, Zongze and Gao, Yue},
  journal={arXiv preprint arXiv:2506.17733},
  year={2025}
}

@article{yang2026m2i2ha,
  title={M2I2HA: A Multi-modal Object Detection Method Based on Intra-and Inter-Modal Hypergraph Attention},
  author={Yang, Xiaofan and Liu, Yubin and Pan, Wei and Chu, Guoqing and Zhang, Junming and Zhao, Jie and Man, Zhuoqi and Cao, Xuanming},
  journal={arXiv preprint arXiv:2601.14776},
  year={2026}
}

@inproceedings{chai2024novel,
  title={A novel adaptive hypergraph neural network for enhancing medical image segmentation},
  author={Chai, Shurong and Jain, Rahul K and Mo, Shaocong and Liu, Jiaqing and Yang, Yulin and Li, Yinhao and Tateyama, Tomoko and Lin, Lanfen and Chen, Yen-Wei},
  booktitle={International Conference on Medical Image Computing and Computer-Assisted Intervention},
  pages={23--33},
  year={2024},
  organization={Springer}
}

@article{jing2025hypergraph,
  title={Hypergraph biformer for semantic segmentation of high-resolution remote sensing images},
  author={Jing, Weipeng and Zhang, Wenjun and Di, Donglin and Li, Chao and Emam, Mahmoud and Mian, Ajmal},
  journal={IEEE Transactions on Geoscience and Remote Sensing},
  year={2025},
  publisher={IEEE}
}

@article{palanisamy2026multi,
  title={Multi-scale attention and hypergraph neural network based deep learning framework for prostate cancer classification and segmentation},
  author={Palanisamy, Manothini and Sennidurai, Karthikumar},
  journal={Biomedical Signal Processing and Control},
  volume={113},
  pages={108925},
  year={2026},
  publisher={Elsevier}
}

@inproceedings{fu2025segman,
  title={SegMAN: Omni-scale context modeling with state space models and local attention for semantic segmentation},
  author={Fu, Yunxiang and Lou, Meng and Yu, Yizhou},
  booktitle={Proceedings of the Computer Vision and Pattern Recognition Conference},
  pages={19077--19087},
  year={2025}
}

@inproceedings{hassani2023neighborhood,
  title={Neighborhood Attention Transformer},
  author={Ali Hassani and Steven Walton and Jiachen Li and Shen Li and Humphrey Shi},
  year=2023,
  booktitle={IEEE/CVF Conference on Computer Vision and Pattern Recognition (CVPR)}
}

@inproceedings{hariharan2011semantic,
  title={Semantic contours from inverse detectors},
  author={Hariharan, Bharath and Arbel{\'a}ez, Pablo and Bourdev, Lubomir and Maji, Subhransu and Malik, Jitendra},
  booktitle={2011 international conference on computer vision},
  pages={991--998},
  year={2011},
  organization={IEEE}
}

@inproceedings{cordts2016cityscapes,
  title={The cityscapes dataset for semantic urban scene understanding},
  author={Cordts, Marius and Omran, Mohamed and Ramos, Sebastian and Rehfeld, Timo and Enzweiler, Markus and Benenson, Rodrigo and Franke, Uwe and Roth, Stefan and Schiele, Bernt},
  booktitle={Proceedings of the IEEE conference on computer vision and pattern recognition},
  pages={3213--3223},
  year={2016}
}

@inproceedings{zhou2017scene,
  title={Scene Parsing through ADE20K Dataset},
  author={Zhou, Bolei and Zhao, Hang and Puig, Xavier and Fidler, Sanja and Barriuso, Adela and Torralba, Antonio},
  booktitle={Proceedings of the IEEE Conference on Computer Vision and Pattern Recognition},
  year={2017}
}

@article{guo2022segnext,
  title={Segnext: Rethinking convolutional attention design for semantic segmentation},
  author={Guo, Meng-Hao and Lu, Cheng-Ze and Hou, Qibin and Liu, Zhengning and Cheng, Ming-Ming and Hu, Shi-Min},
  journal={Advances in neural information processing systems},
  volume={35},
  pages={1140--1156},
  year={2022}
}

@inproceedings{pei2025efficientvmamba,
  title={Efficientvmamba: Atrous selective scan for light weight visual mamba},
  author={Pei, Xiaohuan and Huang, Tao and Xu, Chang},
  booktitle={Proceedings of the AAAI conference on artificial intelligence},
  volume={39},
  number={6},
  pages={6443--6451},
  year={2025}
}

@inproceedings{liu2021swin,
  title={Swin transformer: Hierarchical vision transformer using shifted windows},
  author={Liu, Ze and Lin, Yutong and Cao, Yue and Hu, Han and Wei, Yixuan and Zhang, Zheng and Lin, Stephen and Guo, Baining},
  booktitle={Proceedings of the IEEE/CVF international conference on computer vision},
  pages={10012--10022},
  year={2021}
}

@inproceedings{liu2022convnet,
  title={A convnet for the 2020s},
  author={Liu, Zhuang and Mao, Hanzi and Wu, Chao-Yuan and Feichtenhofer, Christoph and Darrell, Trevor and Xie, Saining},
  booktitle={Proceedings of the IEEE/CVF conference on computer vision and pattern recognition},
  pages={11976--11986},
  year={2022}
}

@article{liu2024vmamba,
  title={Vmamba: Visual state space model},
  author={Liu, Yue and Tian, Yunjie and Zhao, Yuzhong and Yu, Hongtian and Xie, Lingxi and Wang, Yaowei and Ye, Qixiang and Jiao, Jianbin and Liu, Yunfan},
  journal={Advances in neural information processing systems},
  volume={37},
  pages={103031--103063},
  year={2024}
}

@inproceedings{lou2025sparx,
  title={Sparx: A sparse cross-layer connection mechanism for hierarchical vision mamba and transformer networks},
  author={Lou, Meng and Fu, Yunxiang and Yu, Yizhou},
  booktitle={Proceedings of the AAAI Conference on Artificial Intelligence},
  volume={39},
  number={18},
  pages={19104--19114},
  year={2025}
}

@article{zhu2024vision,
  title={Vision mamba: Efficient visual representation learning with bidirectional state space model},
  author={Zhu, Lianghui and Liao, Bencheng and Zhang, Qian and Wang, Xinlong and Liu, Wenyu and Wang, Xinggang},
  journal={arXiv preprint arXiv:2401.09417},
  year={2024}
}

@inproceedings{howard2019searching,
  title={Searching for mobilenetv3},
  author={Howard, Andrew and Sandler, Mark and Chu, Grace and Chen, Liang-Chieh and Chen, Bo and Tan, Mingxing and Wang, Weijun and Zhu, Yukun and Pang, Ruoming and Vasudevan, Vijay and others},
  booktitle={Proceedings of the IEEE/CVF international conference on computer vision},
  pages={1314--1324},
  year={2019}
}

@article{chen2024frequencyaware,
  title={Frequency-aware feature fusion for dense image prediction},
  author={Chen, Linwei and Fu, Ying and Gu, Lin and Yan, Chenggang and Harada, Tatsuya and Huang, Gao},
  journal={IEEE transactions on pattern analysis and machine intelligence},
  volume={46},
  number={12},
  pages={10763--10780},
  year={2024},
  publisher={IEEE}
}

@inproceedings{yu2024embedding,
  title={Embedding-free transformer with inference spatial reduction for efficient semantic segmentation},
  author={Yu, Hyunwoo and Cho, Yubin and Kang, Beoungwoo and Moon, Seunghun and Kong, Kyeongbo and Kang, Suk-Ju},
  booktitle={European Conference on Computer Vision},
  pages={92--110},
  year={2024},
  organization={Springer}
}

@article{yan2024multi,
  title={Multi-scale representations by varying window attention for semantic segmentation},
  author={Yan, Haotian and Wu, Ming and Zhang, Chuang},
  journal={arXiv preprint arXiv:2404.16573},
  year={2024}
}

@inproceedings{yu2018learning,
  title={Learning a discriminative feature network for semantic segmentation},
  author={Yu, Changqian and Wang, Jingbo and Peng, Chao and Gao, Changxin and Yu, Gang and Sang, Nong},
  booktitle={Proceedings of the IEEE conference on computer vision and pattern recognition},
  pages={1857--1866},
  year={2018}
}

@inproceedings{fu2019dual,
  title={Dual attention network for scene segmentation},
  author={Fu, Jun and Liu, Jing and Tian, Haijie and Li, Yong and Bao, Yongjun and Fang, Zhiwei and Lu, Hanqing},
  booktitle={Proceedings of the IEEE/CVF conference on computer vision and pattern recognition},
  pages={3146--3154},
  year={2019}
}

@inproceedings{rahaman2019spectral,
  title={On the spectral bias of neural networks},
  author={Rahaman, Nasim and Baratin, Aristide and Arpit, Devansh and Draxler, Felix and Lin, Min and Hamprecht, Fred and Bengio, Yoshua and Courville, Aaron},
  booktitle={International conference on machine learning},
  pages={5301--5310},
  year={2019},
  organization={PMLR}
}

@article{xu2019frequency,
  title={Frequency principle: Fourier analysis sheds light on deep neural networks},
  author={Xu, Zhi-Qin John and Zhang, Yaoyu and Luo, Tao and Xiao, Yanyang and Ma, Zheng},
  journal={arXiv preprint arXiv:1901.06523},
  year={2019}
}

@inproceedings{magid2021dynamic,
  title={Dynamic high-pass filtering and multi-spectral attention for image super-resolution},
  author={Magid, Salma Abdel and Zhang, Yulun and Wei, Donglai and Jang, Won-Dong and Lin, Zudi and Fu, Yun and Pfister, Hanspeter},
  booktitle={Proceedings of the IEEE/CVF International Conference on Computer Vision},
  pages={4288--4297},
  year={2021}
}

@article{chi2020fast,
  title={Fast fourier convolution},
  author={Chi, Lu and Jiang, Borui and Mu, Yadong},
  journal={Advances in Neural Information Processing Systems},
  volume={33},
  pages={4479--4488},
  year={2020}
}

@article{gu2023adafuse,
  title={AdaFuse: Adaptive medical image fusion based on spatial-frequential cross attention},
  author={Gu, Xianming and Wang, Lihui and Deng, Zeyu and Cao, Ying and Huang, Xingyu and Zhu, Yue-min},
  journal={arXiv preprint arXiv:2310.05462},
  year={2023}
}

@article{canitez2026frequency,
  title={Frequency-Guided Fusion For RGB-Thermal Semantic Segmentation},
  author={Can{\i}tez, {\.I}smail Emre and Erkent, {\"O}zg{\"u}r},
  journal={arXiv preprint arXiv:2605.26273},
  year={2026}
}

@article{han2026fdhnet,
  title={FDHNet: Frequency-aware decomposition with hypergraph modeling network for real-time surface defect segmentation},
  author={Han, Yan and Mu, Yangyong and Huang, Qingqing and Zhang, Yan and Wei, Min and Hong, Seungho},
  journal={Knowledge-Based Systems},
  pages={116853},
  year={2026},
  publisher={Elsevier}
}

\end{document}